\documentclass[10pt,twocolumn,letterpaper]{article}

\usepackage[pagenumbers]{cvpr} 

\usepackage{graphicx}
\usepackage{amsmath}
\usepackage{amssymb}
\usepackage{booktabs}
\usepackage{multirow}
\usepackage{xcolor,colortbl}
\definecolor{LightGray}{RGB}{220, 220, 240}
\definecolor{LightYellow}{RGB}{250, 255, 0}

\definecolor{cvprblue}{rgb}{0.21,0.49,0.74}
\usepackage[pagebackref,breaklinks,colorlinks,allcolors=cvprblue]{hyperref}

\usepackage[capitalize]{cleveref}
\crefname{section}{Sec.}{Secs.}
\Crefname{section}{Section}{Sections}
\Crefname{table}{Table}{Tables}
\crefname{table}{Tab.}{Tabs.}
\usepackage{amsthm}

\newtheorem*{remark}{Remark}
\theoremstyle{definition}
\newtheorem{definition}{Definition}[section]

\def\paperID{****} 
\def\confName{CVPR}
\def\confYear{2025}

\title{Complementary Subspace Low-Rank Adaptation of Vision-Language Models for Few-Shot Classification}

\author{Zhongqi Wang\\
School of EECE, \\University of Chinese Academy of Science\\
Beijing, China\\
{\tt\small wangzhongqi20@mails.ucas.ac.cn}
\and
Jia Dai, Kai Li, Xu Li, Yanmeng Guo \\
Dolby Lab. Inc.\\
Beijing, China\\
{\tt\small \{JiaDai, KaiLi, XuLi, YanmengGuo\}@dolby.com}
\and
Maosheng Xiang\\
School of EECE, \\University of Chinese Academy of Science\\
Beijing, China\\
{\tt\small xms@mail.ie.ac.cn}
}

\begin{document}
\maketitle
\begin{abstract}
    
Vision language model (VLM) has been designed for large scale image-text alignment as a pretrained foundation model. For downstream few shot classification tasks, parameter efficient fine-tuning (PEFT) VLM has gained much popularity in the computer vision community. PEFT methods like prompt tuning and linear adapter have been studied for fine-tuning VLM while low rank adaptation (LoRA) algorithm has rarely been considered for few shot fine-tuning VLM. The main obstacle to use LoRA for few shot fine-tuning is the catastrophic forgetting problem. Because the visual language alignment knowledge is important for the generality in few shot learning, whereas low rank adaptation interferes with the most informative direction of the pretrained weight matrix. We propose the complementary subspace low rank adaptation (Comp-LoRA) method to regularize the catastrophic forgetting problem in few shot VLM finetuning. In detail, we optimize the low rank matrix in the complementary subspace, thus preserving the general vision language alignment ability of VLM when learning the novel few shot information. 
We conduct comparison experiments of the proposed Comp-LoRA method and other PEFT methods on fine-tuning VLM for few shot classification. And we also present the suppression on the catastrophic forgetting problem of our proposed method against directly applying LoRA to VLM. The results show that the proposed method surpasses the baseline method by about +1.0\% Top-1 accuracy and preserves the VLM zero-shot performance over the baseline method by about +1.3\% Top-1 accuracy. The code will be released on github. 
\end{abstract}    

\section{Introduction}

Vision Language Model (VLM) is the most powerful deep learning model for aligning vision and text modals~\cite{clip,align,NEURIPS2022flamingo,ghosh2024exploringfrontiervisionlanguagemodels}. They can even accomplish zero-shot or open-vocabulary tasks. However, with domain shift, VLM may perform poorly on generalizing to unseen datasets. While few data samples are usually accessible, few shot finetuning helps to improve the performance of VLM for new datasets. 

Parameter-efficient fine-tuning (PEFT) methods are widely used to adapt these pre-trained large foundation models for downstream tasks. Therefore, they can be applied to VLM as well. 
Various PEFT methods have been proposed for few shot fine-tuning VLM. CoOp~\cite{coop} firstly applied the prompt tuning method for vision language models. The following works improved prompt tuning VLM for better performance, such as CoCoOp~\cite{cocoop}, KgCoOp~\cite{kgcoop}, PLOT~\cite{plot} and MaPLe~\cite{khattak2023maple}. Another methodology of linear adapter for fine-tuning CLIP has been firstly studied in Clip-Adapter~\cite{clip-adapter}. Then, Tip-Adapter~\cite{tip-adapter}, APE~\cite{Zhu_2023_ICCV_APE}, and many other works move further to incorporate the adapter method for VLM fine-tuning.

While previous works employ various parameter-efficient fine-tuning methods for VLM few-shot classification, the renowned Low Rank Adaptation (LoRA)~\cite{hu2022lora} does not receive much focus in this downstream task as it desired. Only directly using LoRA for few-shot fine-tuning VLM has been demonstrated in~\cite{2024CVPRw_CLIP_LoRA}. 
However, directly using LoRA for few-shot fine-tuning VLM suffers from the catastrophic forgetting problem, as discussed in~\cite{wang2023orthogonal,liang2024inflora,cvpr2024mmaadapter,huisman2023subspaceadaptationpriorfewshot}. 
Few shot fine-tuned vision language models can benefit from the original ability of pretrained model directly. For example, with an extremely limited support dataset, we expect the few shot fine-tuned VLM to generalize to pictures with similar features (same class label). That generalization ability could only be accomplished by the zero-shot knowledge of the pre-trained VLM. 
However, directly applying LoRA may lose this generalization property by overfitting to the few shot support sets. Thus, LoRA fine-tuned VLM performs poorly on the query set with or without the same data distribution. Therefore, LoRA for few-shot fine-tuning CLIP severely suffers from catastrophic forgetting problem.

To regularize the overfitting problem in few-shot fine-tuning VLM via LoRA, we need to gain a comprehensive viewpoint of low rank adaptation. 
LoRA was established on the assumption of the low dimensional property of the parameter space in large foundation models~\cite{li2018measuring,aghajanyan2021intrinsic}. And factorizing the deep neural network has been studied even earlier in~\cite{sainath2013low,zhang2014extracting,povey2018semi}. So LoRA has a mathematical correspondence as principal direction vectors in matrix factorization, for example, the principal singular values and principal singular vectors of SVD decomposition. Our motivation is to regularize the few shot fine-tuning progress via constraining the optimization in the complementary subspace that does not interfere with these principal directions. Most of the pretrained knowledge in these principal directions could be preserved. And the newly learned information of novel classes can be represented in the complementary subspace. 

There are various previous works of regularizing few-shot classification within the transfer-learning framework~\cite{ziko2021laplacianregularizedfewshotlearning,shen2021modelagnosticgraphregularizationfewshot,metalearningsurvey}. For example, entropy regularization~\cite{Dhillon2020Abaselineentropyregularization} enforces the predicted classification logits concentrated intra-class and divergence inter-classes. And we aim to regularize the optimization space of fine-tuning VLM. 
Therefore, our method can be implemented in parallel with other few shot regularization methods, like Shannon entropy regularity~\cite{Dhillon2020Abaselineentropyregularization}, margin maximization~\cite{cvpr2024marginmax}, margin equilibrium~\cite{cvpr2021marginequilibrium}, metric regularization~\cite{2020nipsmetaregularization} etc. 

\textbf{Contributions:} We propose Comp-LoRA, a novel method for fine-tuning VLM to few shot classification via complementary subspace low rank adaptation. 
\begin{enumerate}
    \item Previous parameter-efficient fine-tuning methods on vision language models for few shot learning focus on prompt tuning and adapter-based methods. A recent study on directly using the low rank adaptation method for few shot fine-tuning VLM is limited. We bridge the gap by optimizing the learnable low rank matrix parameters in the complementary subspace for better few shot classification performance. 
    \item This method can suppress the catastrophic forgetting problem for few shot fine-tuning VLM via the complementary subspace restriction. And the proposed method can also be implemented in collaboration with other few shot regularization methods. 
    \item We have done a group of experiments to compare the performance of our proposed method to the previous few shot parameter-efficient fine-tuning methods for VLM. For suppression of the catastrophic forgetting problem, we compare the proposed method with the baseline method on generalizability tasks. We also reveal the effect of complementary subspace dimension to the results through univariate experiments. 
\end{enumerate}

\section{Related Works}
\subsection{Few Shot Classification by Adaptation on VLM}

Few shot classification has been well studied in the area of deep learning and computer vision. Recently, the multi-modal vision language model has been adapted to few shot classification tasks owing to its power to align images and texts.

\textbf{Prompt Tuning}
CoOp~\cite{coop} firstly proposed the prompt tuning method for vision language models. Then, CoCoOp~\cite{cocoop}  and KgCoOp~\cite{kgcoop} both boosted the prompt tuning method by conditional context learning. 
The following work MaPLe~\cite{khattak2023maple} used coupling functions to introduce language tuned features into vision layers for multi-modal prompt tuning. ProGrad~\cite{prograd} projected the interfering gradient to the orthogonal direction. PLOT~\cite{plot} used optimal transport for vision features and fine grain prompt alignment and tuning. PromptSRC~\cite{ICCV2023promptsrc} proposed the self-ensembling strategy to regularize the forgetting problems. 

Recently, CODER~\cite{yic2024coder} and~\cite{cvpr2024quaternion} both enhanced the cross-modal interaction in prompt learning. 
~\cite{icml2024federated_prompt} proposed to use low rank adaptation for prompt modeling in a federated way. And~\cite{chen2024metatuning,cvpr2024domainagnosticprompt,cvpr2024promptmetaregularization} all leveraged the meta-learning framework for prompt tuning. 
Treating the VLM as black box model,~\cite{blackbox} proposed to optimize the prompt via a chat-based LLM. ArGue~\cite{cvpr2024argueattribute},~\cite{cvpr2024anchorrobustfinetuning} and~\cite{cvpr2024tcptextualclass} also used the LLM generated textual prompt for augmentation.~\cite{zhang2024decoupledprompttuning} decoupled the feature channels to maximize the task-shared knowledge.~\cite{cvpr2024anyshift} used a probabilistic graphic model for prompt learning under domain shift. 

\textbf{Adapter Tuning}
Linear adapter for fine-tuning CLIP was firstly studied in ClIP-Adapter~\cite{clip-adapter}. 
In addition to the learning-based methods, Tip-Adapter~\cite{tip-adapter} proposed to use the cache model as anchors for few shot classification, which is a training-free method. The following works, TaskRes~\cite{yu2023task} and APE~\cite{Zhu_2023_ICCV_APE} considered decoupling the prior knowledge and new class knowledge and decided the final classification through hand-crafted rules. 

~\cite{Martin2024transductivefewshotclip} studied the transductive few shot setting of fine-tuning CLIP, and proposed the probabilistic classification method by Dirichlet distribution.~\cite{cvpr2024semanticaided} used the linear adapter as the semantic alignment module for fine-tuning. 

\textbf{Low Rank Adaptation}
Low rank adapter (LoRA) for few-shot fine-tuning vision-Language Models have been studied in~\cite{2024CVPRw_CLIP_LoRA}. The following work MMA~\cite{cvpr2024mmaadapter} only adapted higher layers via multimodal shared LoRA to preserve lower features. And a recent work LLaMP~\cite{Zheng2024LLaMP} combined the LLM knowledge cache guided prompt tuning (for text and vision encoder)  and LoRA (for vision encoder) to get the resulting classification. The subspace concept has also been adopted for few shot learning in~\cite{huisman2023subspaceadaptationpriorfewshot} within the meta training framework. 
Different from these previous works, we propose to regularize the catastrophic forgetting problem via the construction of the parallel adapter module in the complementary subspace.

\begin{figure*}[htb]
    \centering
     \includegraphics[width=.7\linewidth]{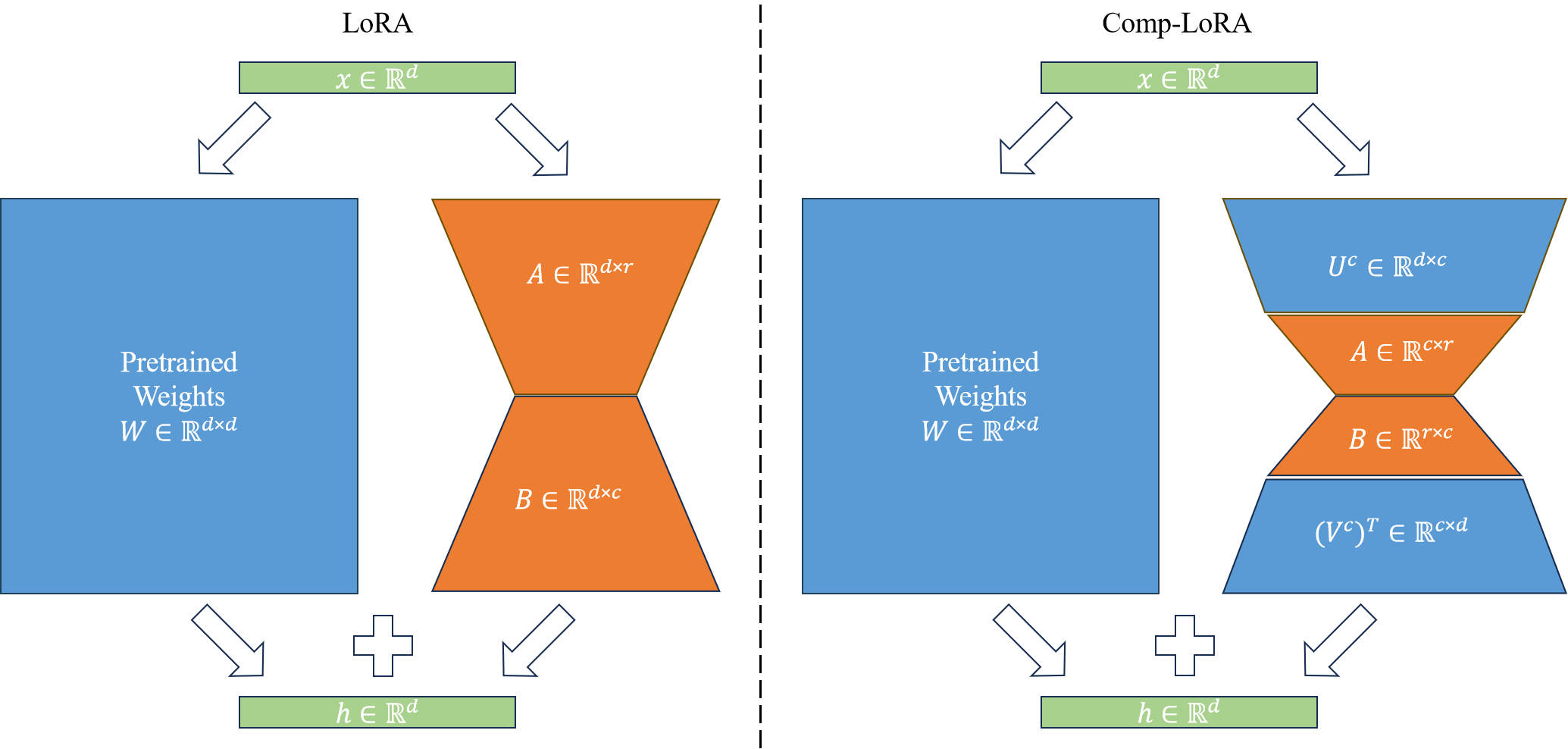}
     \caption{The architecture of Comp-LoRA with comparison to original LoRA.
     We first project the input $x$ into the complementary subspace using the pre-computed matrix $U^c$ and also project the output back to the hidden subspace by matrix $V^c$. In the complementary subspace, we also leverage the LoRA architecture for efficient optimization. }
     \label{fig:comp_lora}
  \end{figure*}
  
\subsection{LoRA Improvements}
With the development of large foundation models, low rank adaptation, as a type of parameter efficient fine tuning method, has attracted more and more attention from the research community. 

\textbf{More Accurate LoRA}
For more accurate fine-tuning comparable with full-fine-tuning, AdaLoRA~\cite{zhang2023adalora} was designed to adaptively optimize the hyper-parameter of intermediate rank by singular value decomposition and sensitivity-based importance scoring. LoRA-GA~\cite{wang2024nips_loraga} used the singular vector of the first step weight update matrix to initialize the low-rank matrices A and B. The following  LoRA-Pro~\cite{wang2024lorapro} modified every update step of LoRA to approximate full fine-tuning. 
Through singular value decomposition, PiSSA~\cite{meng2024pissa} updated the principal components while freezing the residual parts for better performance. 
~\cite{2024olora_QRdecomposition} initialized the low rank matrix using orthonormal vectors through QR decomposition.

\textbf{More Efficient LoRA}
Towards a more efficient learning rate that differs for matrices A and B, LoRA+~\cite{hayou2024ICML_lora_plus} proposed to scale the learning rate of B to be larger than A. 
FourierFT~\cite{icml2024fourierlora} used the Fourier transform basis to approximate the low rank matrix. Similarly, VeRA~\cite{kopiczko2024vera} and NoLA~\cite{iclr2024nola} leveraged random basis for reducing the required parameter amounts. 
A more direct thought is to quantize the large foundation model and~\cite{dettmers2023qlora,kim2023memoryefficientfinetuningcompressedlarge} both combined 4-bit quantization with LoRA to reduce the memory usage. 

\textbf{LoRA and Subspace}
Orthogonal subspace learning to overcome the catestrophy forgetting problem in continual learning has been studied in O-LoRA~\cite{wang2023orthogonal} for LLM and in InfLoRA~\cite{liang2024inflora} for general Foundation models. O-LoRA~\cite{wang2023orthogonal} mitigated catastrophic forgetting of past task knowledge by constraining the gradient updates of the current task to be orthogonal to the gradient of the past tasks. InfLoRA~\cite{liang2024inflora} proposed the interference-free low-rank adaptation (InfLoRA) for continual learning by designing the updating subspace to eliminate the interference between the new and old tasks. 

As mentioned above, singular value decomposition (SVD) has been used in many works~\cite{zhang2023adalora,gamal2023rosasvdupdate,wang2024nips_loraga,meng2024pissa,lingam2024svft}, either for enhancing the performance of LoRA or for reducing the computation resources. Different from them, our method firstly leverages SVD subspace in low rank adapting VLM for few-shot classification.

\section{Methodology} 
Low rank adaptation (LoRA)~\cite{hu2022lora} is a parameter-efficient method for finetuning VLM. When the finetuning data is limited (few shot finetuning), LoRA for VLM suffers from the catastrophic forgetting problem. 
Because the low dimension property of large foundation models may interfere with the optimization direction of LoRA. 
Therefore, we propose to optimize the low rank matrix parameters in the subspace complemented to the principal direction of pretrained weights. 

\subsection{Preliminary: Low Rank Adaptation}
Low rank adaptation (LoRA)~\cite{hu2022lora} has been widely used for fine-tuning large foundation models, such as large language models, visual language models and text-to-image generative models. 
The LoRA module is parallel to the linear weight matrix as two low rank matrix production, presented in the left part of \cref{fig:comp_lora}. 
Mathematically, the linear weight update would be substituted as: 
\begin{align}
  h = W x + \Delta W x = Wx + BAx 
\end{align}

LoRA possesses two highlighted properties of latency-free and parameter-efficiency, owing to the parallel and low rank architecture. Therefore, the computational resources required to fine-tune a large foundation model can be extremely reduced to be conductive for consumer devices. And the storage may be compressed to 100 times the size. So LoRA contributes much to achieve the wide usage of large foundation models for various downstream tasks.

\subsection{Complementary Subspace Low Rank Adaptation}
A mathematical correspondence for LoRA is these principal singular vectors of SVD~\cite{Kanatani2021svdconputervision}. 
And various previous studies have leveraged SVD to initialize LoRA matrices~\cite{meng2024pissa,wang2024nips_loraga} or to optimize the diagonal matrix directly~\cite{gamal2023rosasvdupdate,zhang2023adalora,lingam2024svft}. 
To preserve the vision-text alignment ability of CLIP that is held in principal directions, we need to regularize the VLM few shot fine-tuning process. 
We propose to optimize the low rank matrix in the subspace that is complemented to the principal subspace of pretrained weights. First, we present the rigorous definition of complementary subspace: 

\begin{definition}[Complementary Subspace]
If $\mathbb{R}^d=\mathbb{R}^p\oplus \mathbb{R}^c, d=p+c$ and $\mathbb{R}^p \cap \mathbb{R}^c=0$, then the subspaces $\mathbb{R}^p$ and $\mathbb{R}^c$ are complemented. We call $\mathbb{R}^c$ the complementary subspace, relative to the principal subspace $\mathbb{R}^p$. 
\end{definition}

To get the complementary subspace, we should decompose the weight matrix of linear layers in VLM and figure out the principal directions. 
Here we use the widely adopted singular value decomposition (SVD)~\cite{Kanatani2021svdconputervision} to decompose a linear weight matrix and also obtain the principal scores as follows: 
\begin{align}
    W = U\cdot \Sigma \cdot V, \quad W\in \mathbb{R}^{d\times d}
\end{align}

\begin{figure}[htb]
  \centering
   \includegraphics[width=1.\linewidth]{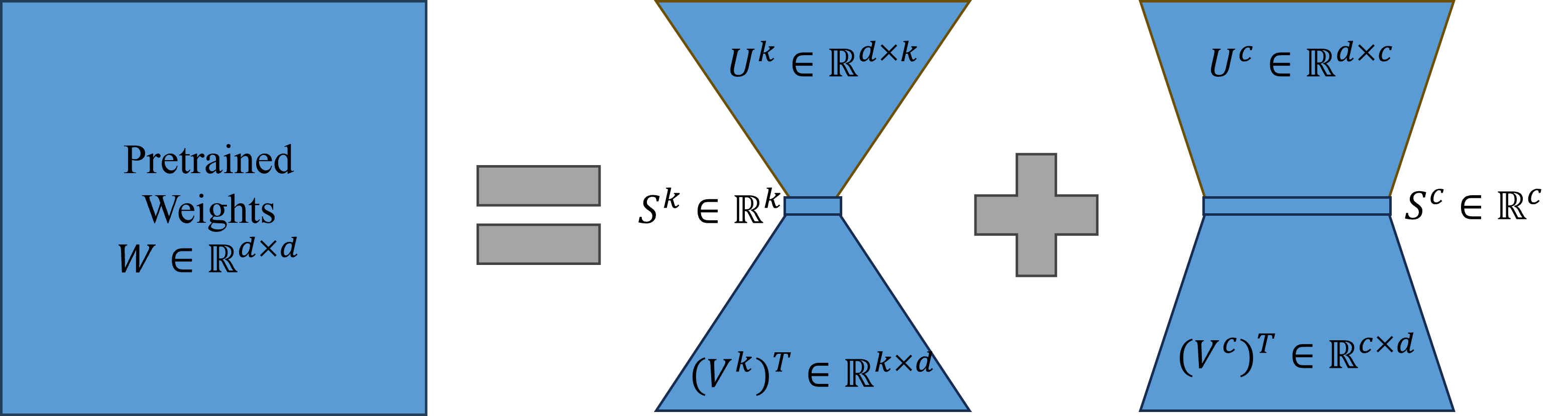}
   \caption{SVD decomposition for weights in linear layers.
   We eliminate the top-k most effective directions and obtain the complementary subspace that is represented by the rest directions.}
   \label{fig:svd_weights}
\end{figure}

With top-p biggest singular values $\Sigma^p$ (principal singular values), the corresponding singular vectors $U^p$ and $V^p$ contribute to the most influence direction of the weight matrix. Intuitively, these singular vectors guide the most changeable directions with the amplification scalars of these singular values in matrix production. 
Then, we eliminate the top-p singular vectors to obtain the complementary subspace through $U^c$ and $V^c$, with the complementary dimension $c=d-p$.  We demonstrate this complementary space decomposition in \cref{fig:svd_weights}. The mathematical formula is: 
\begin{align}
    W = U^p\cdot \Sigma^p \cdot (V^p)^T + U^c\cdot \Sigma^c \cdot (V^c)^T
\end{align} 

\begin{remark}
An extra gain is that SVD results in the orthogonal complementary subspace since the matrices $U$ and $V$ are formed with an orthogonal basis. The orthonormal property further eliminates the interference between principal directions and the update directions, while the complementary property only reduces that interference. 
\end{remark}

\begin{figure}[htb]
  \centering
   \includegraphics[width=.75\linewidth]{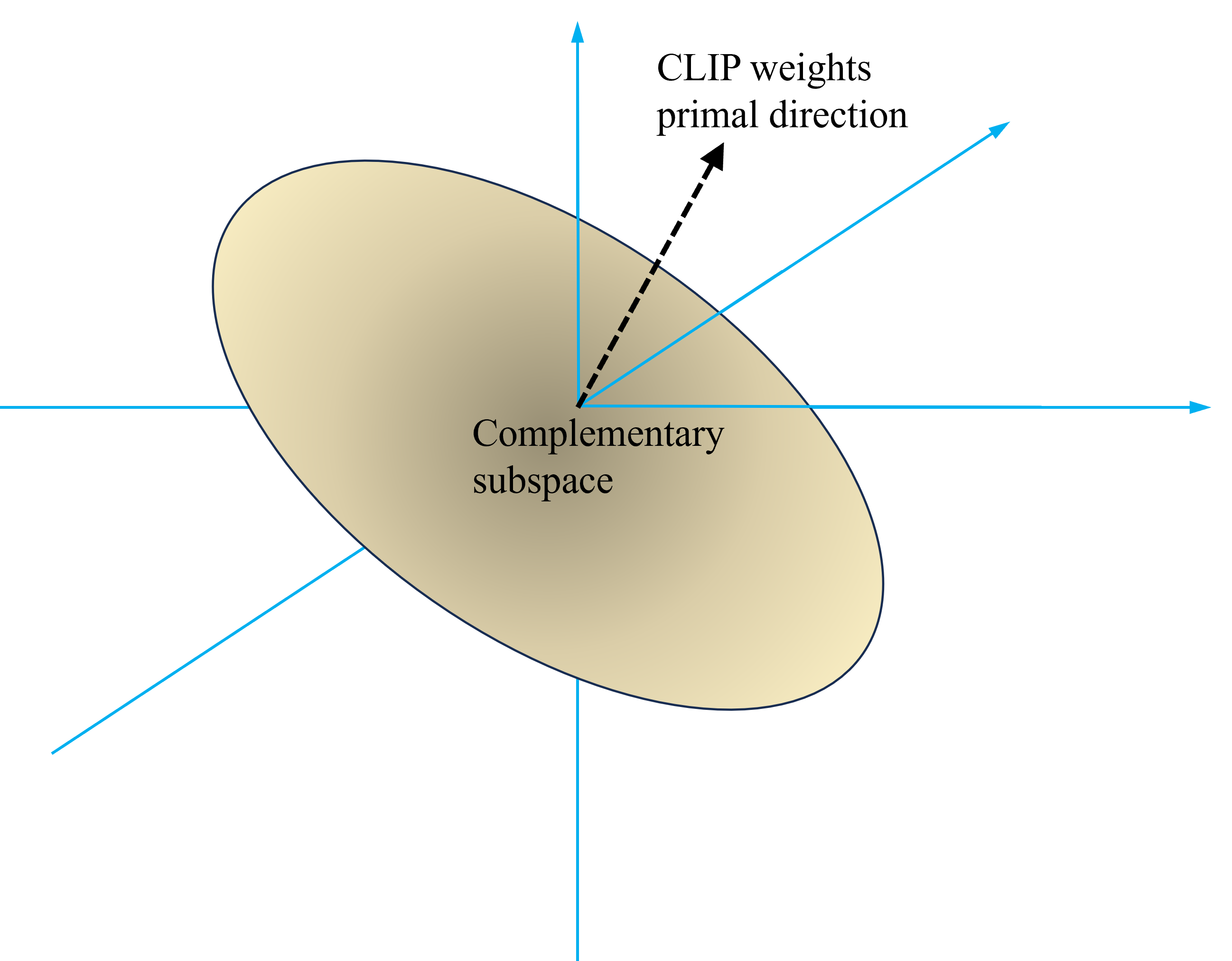}
   \caption{The diagram of complementary subspace.
   We optimize the LoRA module in the subspace that is complemented to the principal directions of the pretrained weight matrix.}
   \label{fig:comp_space}
\end{figure}

After the matrix decomposition, we optimize these learnable parameters in the complementary subspace as shown in \cref{fig:comp_space}. The complementary space does not interfere with the principal directions that hold the zero-shot ability of VLM. And singular value decomposition provides the orthogonal complementary subspace, which prevents the interaction even more strictly. The projection matrix to that complementary subspace is given by the span space of these singular vectors: 
\begin{definition}[Subspace Projection]
    $\mathbb{R}^c$ and $\mathbb{R}^p$ are complemented in $\mathbb{R}^d$. Thus, there exists unique $x\in \mathbb{R}^c$ and $y \in \mathbb{R}^p$ such that $v = x+y$ for every vector $v \in \mathbb{R}^d$. Then the unique linear operator $\mathcal{P} \in \mathbb{C}^{d\times d}$ defined by $\mathcal{P} v = x$ is the projection matrix of $\mathbb{R}^d$ onto $\mathbb{R}^c$ and $x\in \mathbb{R}^c$ is the projection of $v\in \mathbb{R}^d$ onto $\mathbb{R}^c$. Reversely, the pull-back projection $\mathcal{P}^{\dagger}$ retract $x\in \mathbb{R}^c$ onto $v\in \mathbb{R}^d$.
\end{definition}
According to the definition of subspace projection, we assign the projection function $\mathcal{P}$ with the projection matrix $U^c$ and the pull-back projection function $\mathcal{P}^{\dagger}$ with matrix $(V^c)^T$. 

Singular values diagonal matrix $\Sigma^c$ should be discussed. For previous methods~\cite{wang2024nips_loraga,meng2024pissa} that used the principal components of SVD for better initialization, it should be better to incorporate $\Sigma^c$ into the low rank matrix. However, we only need the unitary projection to get the complementary subspace. So the scalar values can be discarded. Besides, the principal singular values are larger than $1$, which makes it effective for the initialization, whereas the complementary singular values even decrease to near zero. These small scales could hinder the optimization progress if adopted into the complementary projection matrix. So we do not incorporate the singular values into the projection matrices. 

In the complementary subspace, the learnable parameters could still be burdensome, since the principal subspace is small while the complementary subspace takes the rest dimensions. So we utilize the low rank matrix factorization method for parameter-efficient learning similar to LoRA, as shown in the right part of \cref{fig:comp_lora}. 
Another factor regarding computational efficiency is the weight matrix decomposition. However it should be conducted only once during the initialization period. And the projection matrix $U^c$ and $(V^c)^T$ are fixed after that. The learnable parameters are the low rank matrix $A$ and $B$. Therefore, the proposed method does not add extra computation during training and inference compared with LoRA. And eliminating the principal directions makes the complementary space slightly smaller than the full weight space. Therefore, the constructed matrices $A$ and $B$ are smaller than those in LoRA module as shown in the right part of \cref{fig:comp_lora}.

\subsection{Comp-LoRA for VLM}\label{sec:comp-lora-for-vlm}
Following the common practice of applying LoRA for large foundation models, we alternatively fine-tune the linear layers of multi-head attention modules in the visual encoder and text encoder of CLIP~\cite{clip}. 
Within the linear layer, we substitute the original linear weight matrix with an extra projected learnable low rank matrix production module in parallelization: 
\begin{align}
    h &= W x + \eta \Delta W x +b \\
    &= W x + \eta (V^c)^T BA U^c x +b
\end{align}
The visual and text encoders in CLIP are constructed by the transformer~\cite{vaswani2017attention} backbone. Each transformer contains $L$  stacked blocks of multi-head attention (MHA) module:
\[
    \text{head}_i=\text{Softmax}\left(\frac{{x}{W}_{q_i} ({x} { W}_{k_i})^T}{\sqrt{d}} \right)({x}{W}_{v_i})
\]
\[
    \text{MHA}({x}) = \text{concat}(\text{head}_1,...,\text{head}_H){ W}_o
\]
For few shot fine-tuning, we should be careful about the choice of these linear layers to be substituted. 
However, this choice should be independent of the Comp-LoRA algorithm. Various works~\cite{align,cvpr2024mmaadapter} have discussed the position choice to be substituted. With the same layers in the model substituted, the performance gap in the comparison experiments should reflect the difference between Comp-LoRA and these baseline methods only. 

\subsection{Optimization Property}\label{sec:optimization_property}
Then, whether the proposed Comp-LoRA method converges to the desired optimal point or not? Briefly, the optimization occurs in a rotated and dimension reduced subspace to the contrary of full space in vanilla LoRA. Rotation hardly affects the optimization process. Whereas dimension reduction may result in the suboptimal solution in the full space, which is the optimal solution in the subspace. 

The optimization follows a gradient-based scheme. 
Recall the gradients of LoRA~\cite{wang2024nips_loraga}: 
\begin{align}
    W &= W_0 + \Delta W = W_0 + BA, \\
    B_{t+1} &= B_{t} + \eta (\Delta W -BA)A_{t} \\
    A_{t+1} &= A_{t} + \eta (\Delta W -BA)B_{t} 
\end{align}
The gradients are rotated by the projection matrix spanned by all singular vectors: 
\begin{align}
  W &= W_0 + \Delta W = W_0 + V^TBAU, \\
  B_{t+1} &= B_{t} + \eta (\Delta W -V^TBAU)V^TA_{t}U  \\
  A_{t+1} &= A_{t} + \eta (\Delta W -V^TBAU)V^TB_{t}U  
\end{align}
The $U$ and $V$ matrix above denote the full rank projection of singular value decomposition. 
Then, to suppress the interference with pre-trained VLM, we propose to optimize the learnable low rank matrix $A$ and $B$ in the complementary subspace. 
\begin{align}
  W &= W_0 + \Delta W = W_0 + (V^c)^TBAU^c, \\
  B_{t+1} &= B_{t} + \eta (\Delta W -(V^c)^TBAU^c)(V^c)^T A_{t}U^c  \\
  A_{t+1} &= A_{t} + \eta (\Delta W -(V^c)^TBAU^c)(V^c)^T B_{t}U^c  
\end{align}
So the transformation should incorporate the dimension reduction process as well. One interpretation is that the rotated gradients are projected to the subspace, resulting in the projected gradient descent~\cite{Boyd_Vandenberghe_2004Convex_Optimization}. 

Another intuitive way to understand this subspace optimization process is to reformulate the objective function with scales. 
\begin{align}
  & \min_{A,B} \|(V^c)^TBAU^c - \Delta W\|^2_{F} \\
  \Rightarrow \quad & \min_{A,B} \|BA - (V^c)\Delta W (U^c)^T\|^2_{F} 
\end{align}
which is equivalent to approximating the projected weight update matrix with low rank matrix production.

\begin{table*}[t!]
\caption{Comparison experimental results on 11 few shot classification tasks. We averaged over 5 random seeds for the Top-1 accuracy values. The highest value is highlighted in \textbf{bold}, and the second-highest is \underline{underlined}.}
\label{tab:few_shot_datasets}
\centering
\resizebox{\textwidth}{!}{
\renewcommand{\arraystretch}{1.06} 
\begin{tabular}{llcccccccccccc}
\toprule
Shots & Method & ImageNet & SUN & Aircraft & EuroSAT & Cars & Food & Pets &  Flowers & Caltech & DTD & UCF & Average
\\ \midrule 
\multirow{1}{*}{0} & \textbf{CLIP} {\tiny \textbf{(ICML '21)}} & 66.7 & 62.6 & 24.7 & 47.5 & 65.3 & 86.1 & 89.1 & 71.4 & 92.9 & 43.6 & 66.7 & 65.1 \\
\midrule
\midrule
\multirow{11}{*}{\textbf{1}} 
& CoOp {\tiny \textbf{(IJCV '22)}} & 68.0 & 67.3 & 26.2 & 50.9 & 67.1 & 82.6 & 90.3 & 72.7 & 93.2 & 50.1 & 70.7 & 67.2 \\
& CoCoOp {\tiny \textbf{(CVPR '22)}} & 69.4 & 68.7 & 28.1 & 55.4 & 67.6 & 84.9 & 91.9 & 73.4 & 94.1 & 52.6 & 70.4 & 68.8 \\
& TIP-Adapter-F {\tiny \textbf{(ECCV '22)}} & 69.4 & 67.2 & 28.8 & {67.8} & 67.1 & 85.8 & 90.6 & \underline{83.8} & 94.0 & 51.6 & 73.4 & {70.9} \\
& CLIP-Adapter {\tiny \textbf{(IJCV '23)}} & 67.9 & 65.4 & 25.2 & 49.3 & 65.7 & 86.1 & 89.0 & 71.3 & 92.0 & 44.2 & 66.9 & 65.7 \\
& PLOT++ {\tiny \textbf{(ICLR '23)}} & 66.5 & 66.8 & 28.6 & 65.4 & {68.8} & \underline{86.2} & 91.9 & 80.5 & \underline{94.3} & \underline{54.6} & {74.3} & 70.7 \\
& KgCoOp {\tiny \textbf{(CVPR '23)}} & 68.9 & 68.4 & 26.8 & 61.9 & 66.7 & \textbf{86.4} & \underline{92.1} & 74.7 & {94.2} & 52.7 & 72.8 & 69.6 \\
& TaskRes {\tiny \textbf{(CVPR '23)}} & 69.6 & 68.1 & \textbf{31.3} & 65.4 & {68.8} & 84.6 & 90.2 & 81.7 & 93.6 & 53.8 & 71.7 & 70.8 \\
& MaPLe {\tiny \textbf{(CVPR '23)}} & {69.7} & {69.3} & 28.1 & 29.1 & 67.6 & 85.4 & 91.4 & 74.9 & 93.6 & 50.0 & 71.1 & 66.4 \\
& ProGrad {\tiny \textbf{(ICCV '23)}} & 67.0 & 67.0 & 28.8 & 57.0 & 68.2 & 84.9 & 91.4 & 80.9 & 93.5 & 52.8 & 73.3 & 69.5  \\
& APE {\tiny \textbf{(ICCV '23)}} & 70.29& 69.78& 30.48& 65.16& 68.98& 85.91& 90.00& 88.71& 94.69& 56.56& 72.35  & 72.08 \\
& CLIP-LoRA {\tiny \textbf{(CVPRW '24)}}& \textbf{70.4} & \textbf{70.4} & \underline{30.2} & \underline{72.3} & \textbf{70.1} & 84.3 & \textbf{92.3} & {83.2} & 93.7 & {54.3} & \underline{76.3} & \underline{72.5} \\
\rowcolor{LightGray} & Comp-LoRA (Ours) & \underline{69.97} & \underline{70.09} & {29.81} & \textbf{79.93} & \underline{69.84} & 84.40 & {91.62} & \textbf{85.38} & \textbf{94.52} & \textbf{59.16} & \textbf{77.72} & \textbf{73.85} \\
\midrule
\midrule
\multirow{11}{*}{\textbf{4}}
& CoOp  {\tiny \textbf{(IJCV '22)}} &  69.7 & 70.6 & 29.7 & 65.8 & 73.4 & 83.5 & 92.3 & 86.6 & 94.5 & 58.5 & 78.1 & 73.0 \\
& CoCoOp {\tiny \textbf{(CVPR '22)}} & 70.6 & 70.4 & 30.6 & 61.7 & 69.5 & 86.3 & \underline{92.7} & 81.5 & 94.8 & 55.7 & 75.3 & 71.7 \\
& TIP-Adapter-F {\tiny \textbf{(ECCV '22)}} & 70.7 & 70.8 & {35.7} & 76.8 & 74.1 & 86.5 & 91.9 & 92.1 & 94.8 & 59.8 & 78.1 & 75.6 \\
& CLIP-Adapter {\tiny \textbf{(IJCV '23)}} & 68.6 & 68.0 & 27.9 & 51.2 & 67.5 & 86.5 & 90.8 & 73.1 & 94.0 & 46.1 & 70.6 & 67.7 \\
& PLOT++ {\tiny \textbf{(ICLR '23)}} & 70.4 & 71.7 & 35.3 & {83.2} & {76.3} & 86.5 & 92.6 & {92.9} & {95.1} & {62.4} & {79.8} & {76.9} \\
& KgCoOp {\tiny \textbf{(CVPR '23)}} & 69.9 & 71.5 & 32.2 & 71.8 & 69.5 & \textbf{86.9} & 92.6 & 87.0 & 95.0 & 58.7 & 77.6 & 73.9 \\
& TaskRes {\tiny \textbf{(CVPR '23)}} & \underline{71.0} &{72.7} & 33.4 & 74.2 & 76.0 & 86.0 & 91.9 & 85.0 & 95.0 & 60.1 & 76.2 & 74.7 \\
& MaPLe {\tiny \textbf{(CVPR '23)}} & 70.6 & 71.4 & 30.1 & 69.9 & 70.1 & \underline{86.7} & \textbf{93.3} & 84.9 & 95.0 & 59.0 & 77.1 & 73.5  \\
& ProGrad {\tiny \textbf{(ICCV '23)}} & 70.2 & 71.7 & 34.1 & 69.6 & 75.0 & 85.4 & 92.1 & 91.1 & 94.4 & 59.7 & 77.9 & 74.7 \\
& APE {\tiny \textbf{(ICCV '23)}} & 70.80& 72.36& 34.68& 75.77& 73.36& 86.27& 91.58& 94.64& 95.58& 65.54& 78.85   & 76.31 \\
& CLIP-LoRA {\tiny \textbf{(CVPRW '24)}}& \textbf{71.4} &  \underline{72.8} &  \underline{37.9} & \underline{84.9} & \textbf{77.4} & 82.7 & 91.0 & \underline{93.7} & \underline{95.2} & \underline{63.8} & \textbf{81.1} & \underline{77.4} \\
\rowcolor{LightGray} & Comp-LoRA (Ours) & \textbf{71.4} & \textbf{73.11} & \textbf{38.32} & \textbf{86.4} & \underline{76.73} & 82.7 & 90.29 & \textbf{94.03} & \textbf{95.28} & \textbf{64.54} & \underline{80.97} & \textbf{77.61} \\
\midrule
\midrule
\multirow{11}{*}{\textbf{16}}
& CoOp  {\tiny \textbf{(IJCV '22)}} &  71.5 & 74.6 & 40.1 & 83.5 & 79.1 & 85.1 & 92.4 & 96.4 & 95.5 & 69.2 & 81.9 & 79.0 \\
& CoCoOp {\tiny \textbf{(CVPR '22)}} & 71.1 & 72.6 & 33.3 & 73.6 & 72.3 & \textbf{87.4} & \underline{93.4} & 89.1 & 95.1 & 63.7 & 77.2 & 75.4 \\
& TIP-Adapter-F {\tiny \textbf{(ECCV '22)}} & {73.4} &  {76.0} & 44.6 & 85.9 & 82.3 & 86.8 & 92.6 & 96.2 & 95.7 & 70.8 & 83.9 & 80.7 \\
& CLIP-Adapter {\tiny \textbf{(IJCV '23)}} & 69.8 & 74.2 & 34.2 & 71.4 & 74.0 & 87.1 & 92.3 & 92.9 & 94.9 & 59.4 & 80.2 & 75.5  \\
& PLOT++ {\tiny \textbf{(ICLR '23)}} & 72.6 & {76.0} & {46.7} & {92.0} & {84.6} & 87.1 & \textbf{93.6} & {97.6} & {96.0} & 71.4 & {85.3} & {82.1} \\
& KgCoOp {\tiny \textbf{(CVPR '23)}} & 70.4 & 73.3 & 36.5 & 76.2 & 74.8 & \underline{87.2} & 93.2 & 93.4 & 95.2 & 68.7 & 81.7 & 77.3 \\
& TaskRes {\tiny \textbf{(CVPR '23)}} & 73.0 & \underline{76.1} & 44.9 & 82.7 & 83.5 & 86.9 & 92.4 & 97.5 & 95.8 & {71.5} & 84.0 & 80.8 \\
& MaPLe {\tiny \textbf{(CVPR '23)}} &  71.9 & 74.5 & 36.8 & 87.5 & 74.3 & \textbf{87.4} & 93.2 & 94.2 & 95.4 & 68.4 & 81.4 & 78.6 \\
& ProGrad {\tiny \textbf{(ICCV '23)}} &  72.1 & 75.1 & 43.0 & 83.6 & 82.9 & 85.8 & 92.8 & 96.6 & 95.9 & 68.8 & 82.7 & 79.9 \\
& APE {\tiny \textbf{(ICCV '23)}} & 71.48& 74.22 &42.63& 81.57& 77.19& 86.72& 92.01& 94.84& 95.38& 69.98& 80.76  & 78.79 \\
& CLIP-LoRA {\tiny \textbf{(CVPRW '24)}}& \underline{73.6} & \underline{76.1} & \underline{54.7} & \underline{92.1}  & \underline{86.3} & 84.2 & 92.4 & \textbf{98.0} & \underline{96.4} & \underline{72.0} & \textbf{86.7} & \underline{83.0} \\
\rowcolor{LightGray} & Comp-LoRA (Ours) & \textbf{73.72} & \textbf{76.52} & \textbf{56.53} & \textbf{93.25}  & \textbf{87.1} & 84.21 & 93.14 & \underline{97.97} & \textbf{96.80} & \textbf{72.12} & \underline{86.62} & \textbf{83.45} \\
\bottomrule

\end{tabular}}

\end{table*}

\section{Experiments}

\subsection{Experimental Setup}
\textbf{Datasets}
We follow the setting of previous few shot classification studies~\cite{2024CVPRw_CLIP_LoRA,coop,tip-adapter,Zhu_2023_ICCV_APE}.
There are 11 datasets for fine-grained classification: scenes (SUN397~\cite{sun397}), aircraft types (Aircraft~\cite{aircraft}), remote sensing (EuroSAT~\cite{eurosat}), automobiles (Stanford-Cars~\cite{cars}), food items (Food101~\cite{food}), pet breeds (Oxford-Pets~\cite{pets}), flowers (Flower102~\cite{flower}), general objects (Caltech101~\cite{caltech101}), textures (DTD~\cite{dtd}) and human actions (UCF101~\cite{ucf101}) as well as ImageNet~\cite{imagenet}. These datasets cover most scenarios and are capable of forming a thorough benchmark for few shot visual classification. 

\textbf{Training}
We use the pretrained CLIP method with the vision transformer backbone for LoRA implementation. We adopt the ViT-B/16 backbone for these potential linear layers that could be parallel with the Comp-LoRA module. The most important hyper-parameter of LoRA is the low rank dimension $r$. For few shot fine-tuning VLM, the low rank dimension should be small given the limited amount of support data pairs. For the conveninence of comparison, we follow the setting in the baseline work~\cite{2024CVPRw_CLIP_LoRA} as $r=2$. The initial learning rate is set to $2^{-4}$. 
For few shot training, we construct the support set according to the n-shot settings in different experiments. 

\subsection{Comparison Experiments on Fine-tuning VLM for Few Shot Classification}
\textbf{Baseline} 
We compare the proposed Comp-LoRA method with several previous methods. For prompt tuning methods, we choose the typical methods~\cite{coop,kgcoop} and state-of-the-art methods~\cite{plot,prograd,khattak2023maple} in this series for comparison.  
For adapter-based methods, we choose the most influencing works of CLIP-Adapter~\cite{clip-adapter}, Tip-Adapter~\cite{tip-adapter}, TaskRes~\cite{yu2023task} and APE~\cite{Zhu_2023_ICCV_APE} for comparing. 
Some other methods~\cite{Zheng2024LLaMP,cvpr2024mmaadapter} can be implemented with our algorithm as a collaborative solution for few shot classification, that are not included in our comparison experiments. 

\begin{figure}[htb]
    \centering
    \includegraphics[width=.85\linewidth]{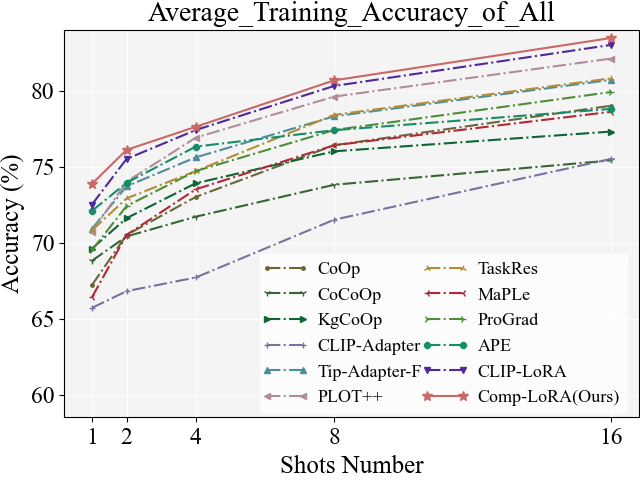}
    \caption{The comparison experiments of different methods on the average accuracy score.
    The proposed method Comp-LoRA outperforms other methods.}
    \label{fig:average_acc_all}
\end{figure}

\textbf{Results}
The results of 1-shot, 4-shot and 16-shot experiments are demonstrated in \cref{tab:few_shot_datasets}. Most Comp-LoRA results achieve the highest or the second-highest accuracy. 
For better visualization, we plot the accuracy curves of the average scores over 11 datasets. As shown in \cref{fig:average_acc_all}, our proposed method presents the highest average performance. 
For the complete results of other n-shot settings, please refer to the supplementary materials. 

Note that with extremely limited support data as shown in the 1-shot situation, the LoRA-type methods behave poorly. Whereas prompt tuning methods (PLOT, KgCoOp) and prior-based methods (Tip-Adapter, APE) behave slightly better. Because LoRA-type methods need to learn the adaptation knowledge from scratch while prompt tuning methods and prior-based methods leverage pretty much human knowledge into the classification process. 

Two datasets (Food, Pets) present inconsistent results compared with others, on which the prompt-tuning methods performs better. As stated in~\cite{kaur2017combiningweaklyweblysupervised}, there are noise labels in the training set of Food101. Similarly, the images of Oxford-Pets dataset have a large variations in scale, pose and lighting~\cite{catsanddogs}. Further study is expected to improve upon these two datasets.

\subsection{Suppression on Catastrophic Forgetting Problem}

To evaluate the performance of our proposed method for suppressing the catastrophic forgetting problem, a good choice is to test the preserved ability of CLIP after fine-tuning. So we fine-tune CLIP on one few shot support set and then test the zero-shot ability of the fine-tuned CLIP on the other few shot query sets. As shown in \cref{tab:leave_one_out}, we fine-tune with the ImageNet support set and test on the other 10 few shot tasks similar to cross-validation. For few shot learning, the proposed Comp-LoRA method performs well in the fine-tuning tasks and gets a better accuracy score than the baseline CLIP-LoRA method on the ImageNet few shot task. At the same time, Comp-LoRA outperforms the baseline CLIP-LoRA method on most other zero-shot tests. Note that there is a performance gap between the zero shot performance of fine-tuned models and pretrained CLIP without finetuning. 
We further conduct more experiments on different datasets as the training set. Please find these extra experiments in the supplementary materials. 
\begin{table*}[t!]
\caption{Designed experiments on the catastrophic forgetting problem. We fine-tune on ImageNet support set through the proposed and baseline methods, and then test their reserved zero shot classification ability on other 10 few shot tasks. The average column was taken on the other 10 few shot datasets. We set the backbone as ViT-B/16. The complementary subspace dimension in Comp-LoRA was set as 496. We average over 5 random seeds for the Top-1 accuracy values. The highest value is highlighted in \textbf{bold}.}
\label{tab:leave_one_out}
\centering
\resizebox{\textwidth}{!}{
\renewcommand{\arraystretch}{1.06} 
\begin{tabular}{l|c|ccccccccccc}
\toprule
Method &  ImageNet  & SUN & Aircraft & EuroSAT & Cars & Food & Pets &  Flowers & Caltech & DTD & UCF & Average
\\ \midrule 
{CLIP} {\tiny \textbf{(ICML '21)}} & 66.7 & 62.6 & 24.7 & 47.5 & 65.3 & 86.1 & 89.1 & 71.4 & 92.9 & 43.6 & 66.7 & 64.99 \\
\midrule
CoOp {\tiny \textbf{(IJCV '22)}}& {71.51} & {64.15} & {18.47} & {46.39} & { 64.51} &  85.30 & {89.14} & {68.71} &  93.70 & { 41.92} & {66.55 } & {63.88} \\
CoCoOp {\tiny \textbf{(CVPR '22)}}& {71.02} & { 67.36} & {22.94} & {45.37} & {65.32} & 86.06 & {90.14} & {71.88} &  \textbf{94.43} & { 45.73} & { 68.21} & {65.74} \\
MaPLe {\tiny \textbf{(CVPR '23)}}& {70.72} & {67.01} & \textbf{24.74} & \textbf{48.06} & {65.57} & {86.20} & {90.49} & {72.23} &  93.53  & {46.49} & {68.69} & {66.30} \\
PromptSRC {\tiny \textbf{(ICCV '23)}}& {71.27} & {67.10} & {23.90} & {45.50} & {65.70} &  86.15 & {90.25} & {70.25} &  93.60 & {46.87} & {68.75 } & {65.81} \\
MMA {\tiny \textbf{(CVPR '24)}}& {71.00} & {68.17} & {25.33} & {46.57} & {66.13} &  {86.12} & {90.30} & {72.07} &  93.80 & {46.57} & {68.32 } & {66.61} \\
    CLIP-LoRA {\tiny \textbf{(CVPRW '24)}}& \textbf{73.42} & 67.44 & 23.67 & 45.70 & 64.30 & 85.79 & 89.18 & 71.00 & 93.79 & 44.80 & 67.35 & {65.30} \\
Comp-LoRA (Ours) & {73.35} & \textbf{68.82} & {24.57} & {47.90} & \textbf{66.59} & \textbf{86.68} & \textbf{90.62} & \textbf{72.82} & {94.08} & \textbf{47.16} & \textbf{68.81} & \textbf{66.80} \\
\bottomrule
\end{tabular}}
\end{table*}

\subsection{The Effect of Complementary Subspace Dimension}
In our proposed complementary subspace low rank adaptation method, the hyper-parameter of the complementary subspace dimension plays a key role on the final performance. We conduct a group of univariate experiments to show the effect of choice on the complementary subspace dimension. 

\begin{figure}[htb]
    \centering
    \includegraphics[width=.85\linewidth]{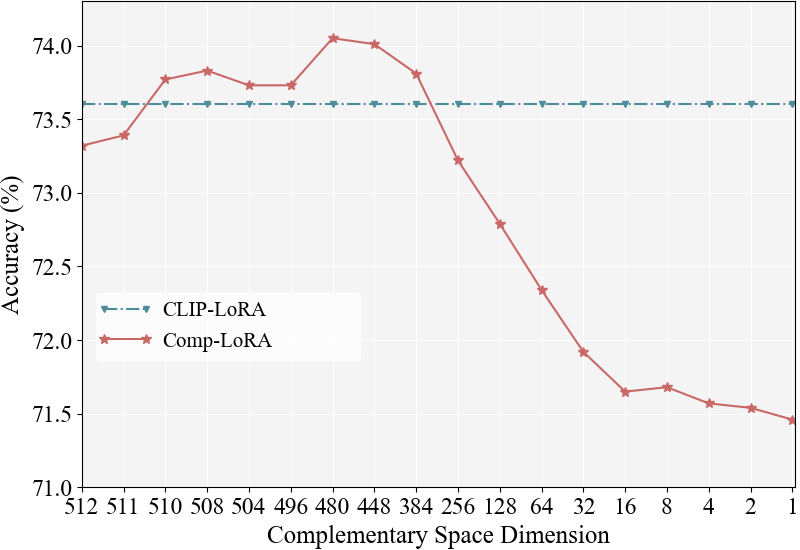}
    \caption{The effect of complementary subspace dimension on ImageNet. 
    The X-axis is scaled for better demonstration. 
    The performance of the proposed Comp-LoRA firstly increases and then decreases when the complementary subspace dimension decreases, as far as the main trend. Within the range of $[511, 384]$, the proposed Comp-LoRA outperforms the baseline CLIP-LoRA method.}
    \label{fig:average_comp_rank}
\end{figure}
We choose the typical dataset \emph{ImageNet} for demonstration, as shown in \cref{fig:average_comp_rank}.
For the supplementary subspace dimension change schedule, we use the exponential law of $2^n, n\in\{0, 1, \cdots, 8\}$ and reverse them for the second half, then subtract the full dimension for the first half, as the X-axis in \cref{fig:average_comp_rank}. 
The performance of the proposed Comp-LoRA method increases at the beginning and then decreases when the complementary subspace dimension decreases. 
A possible explanation is that there is a trade-off between the suppression of catastrophic forgetting and efficient adaptation. The full space optimization does not suppress the catastrophic forgetting of VLM, thus causing poor evaluation results at the beginning of the curve. When we begin to eliminate the principle directions, both suppression of catastrophic forgetting and efficient fine-tuning are achievable. 
Therefore, within the range of $[511, 384]$, our proposed method surpasses the baseline method.

\begin{figure}[htb]
    \centering
    \includegraphics[width=.75\linewidth]{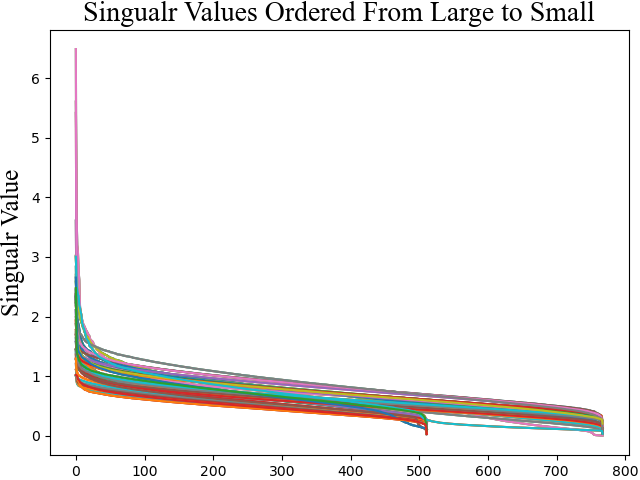}
    \caption{The singular values of all linear weight matrix in pretrained CLIP.
    We demonstrate the results in the arranged order from large to small. These prominent singular values mostly stay in the first 16 dimensions.}
    \label{fig:sfull}
\end{figure}

For choosing the optimal complementary subspace dimension, we need to combine the relative performance and the robustness. Therefore, $496$ is a good choice with a relatively highest accuracy score and middle position within the range of $[511, 384]$. The complementary subspace dimension hyper-parameter should be smaller than the computed decomposed complementary subspace of the pretrained weight matrix in VLM. We provide the singular value distribution of all linear weight matrices in a pretrained CLIP model, as shown in \cref{fig:sfull}. Most principal singular values stay in the first $16$ dimensions approximately. So the univariate experiments on the complementary subspace dimension coincide with the distribution of singular values, since these linear projection matrices in multi-head attention are $512$ dimensions, with the $16$ principal dimensions and $496$ complementary dimensions. 

When the complementary subspace dimension decreases down a certain value, the efficient fine-tuning ability should lost. Because the squeezed complementary subspace only allows limited optimization directions. In the reduced subspace, the optimization may be affected and only suboptimal solution is achievable, as discussed in \cref{sec:optimization_property}.
We also conduct univariate experiments on other datasets. 
Please find these experimental results in the supplementary materials.

\section{Conclusion} 
We studied the parameter-efficient fine-tuning method of vision language model for few shot classification. To suppress the catastrophic forgetting problem of directly applying LoRA for VLM, we proposed to decompose the weight matrix space and optimize the low rank adaptation module in the complementary subspace. We provided a thorough illustration of the computation of matrix space decomposition and the optimization property in the complementary subspace. 
Next, we conducted comparison experiments of our proposed Comp-LoRA method and previous parameter-efficient fine-tuning methods on various few shot datasets. The results showed that Comp-LoRA surpassed other methods. Furthermore, we designed the experiments to demonstrate the suppression of the catastrophic forgetting problem of our proposed method. Finally, we did univariate experiments on the dimension hyper-parameter of complementary subspace. Among a certain range, our proposed method performs better than the baseline method, which suggests a proper choice of the dimension hyper-parameter. 

A future study may focus on the possibly more advanced parameter-efficient fine-tuning vision language model methods for few shot classification. Another attractive direction is multi-task learning by applying LoRA to VLM given the low storage of LoRA. Regarding theory, we still need a solid explanation for the superior performance of low rank adaptation over other PEFT methods.

{
    \small
    \bibliographystyle{ieeenat_fullname}
    \bibliography{main}
}

\newpage 
\setcounter{page}{1}
\maketitlesupplementary

\section{Complete Few Shot Experiments}
\label{sec:rationale}
We present the complete experimental results of \{1,2,4,8,16\}-shots in \cref{tab:few_shot_full}. On most of these few shot datasets, our Comp-LoRA method achieves the highest or second highest performance. 

\begin{table*}[h!]
\caption{Comparison experimental results on 11 few shot classification tasks with the ViT-B/16, CLIP. We averaged over 5 random seeds for the Top-1 accuracy values. Highest value is highlighted in \textbf{bold}, and the second highest is \underline{underlined}.}
\label{tab:few_shot_full}
\centering
\resizebox{.97\textwidth}{!}{
\begin{tabular}{llcccccccccccc}
\toprule
Shots & Method & ImageNet & SUN & Aircraft & EuroSAT & Cars & Food & Pets &  Flowers & Caltech & DTD & UCF & Average
\\ \midrule 
\multirow{1}{*}{0} & \textbf{CLIP} {\tiny \textbf{(ICML '21)}} & 66.7 & 62.6 & 24.7 & 47.5 & 65.3 & 86.1 & 89.1 & 71.4 & 92.9 & 43.6 & 66.7 & 65.1 \\
\midrule
\midrule
\multirow{11}{*}{\textbf{1}} 
& CoOp {\tiny \textbf{(IJCV '22)}} & 68.0 & 67.3 & 26.2 & 50.9 & 67.1 & 82.6 & 90.3 & 72.7 & 93.2 & 50.1 & 70.7 & 67.2 \\
& CoCoOp {\tiny \textbf{(CVPR '22)}} & 69.4 & 68.7 & 28.1 & 55.4 & 67.6 & 84.9 & 91.9 & 73.4 & 94.1 & 52.6 & 70.4 & 68.8 \\
& TIP-Adapter-F {\tiny \textbf{(ECCV '22)}} & 69.4 & 67.2 & 28.8 & {67.8} & 67.1 & 85.8 & 90.6 & \underline{83.8} & 94.0 & 51.6 & 73.4 & {70.9} \\
& CLIP-Adapter {\tiny \textbf{(IJCV '23)}} & 67.9 & 65.4 & 25.2 & 49.3 & 65.7 & 86.1 & 89.0 & 71.3 & 92.0 & 44.2 & 66.9 & 65.7 \\
& PLOT++ {\tiny \textbf{(ICLR '23)}} & 66.5 & 66.8 & 28.6 & 65.4 & {68.8} & \underline{86.2} & 91.9 & 80.5 & \underline{94.3} & \underline{54.6} & {74.3} & 70.7 \\
& KgCoOp {\tiny \textbf{(CVPR '23)}} & 68.9 & 68.4 & 26.8 & 61.9 & 66.7 & \textbf{86.4} & \underline{92.1} & 74.7 & {94.2} & 52.7 & 72.8 & 69.6 \\
& TaskRes {\tiny \textbf{(CVPR '23)}} & 69.6 & 68.1 & \textbf{31.3} & 65.4 & {68.8} & 84.6 & 90.2 & 81.7 & 93.6 & 53.8 & 71.7 & 70.8 \\
& MaPLe {\tiny \textbf{(CVPR '23)}} & {69.7} & {69.3} & 28.1 & 29.1 & 67.6 & 85.4 & 91.4 & 74.9 & 93.6 & 50.0 & 71.1 & 66.4 \\
& ProGrad {\tiny \textbf{(ICCV '23)}} & 67.0 & 67.0 & 28.8 & 57.0 & 68.2 & 84.9 & 91.4 & 80.9 & 93.5 & 52.8 & 73.3 & 69.5  \\
& APE {\tiny \textbf{(ICCV '23)}} & 70.29& 69.78& 30.48& 65.16& 68.98& 85.91& 90.00& 88.71& 94.69& 56.56& 72.35  & 72.08 \\
& CLIP-LoRA {\tiny \textbf{(CVPRW '24)}}& \textbf{70.4} & \textbf{70.4} & \underline{30.2} & \underline{72.3} & \textbf{70.1} & 84.3 & \textbf{92.3} & {83.2} & 93.7 & {54.3} & \underline{76.3} & \underline{72.5} \\
\rowcolor{LightGray} & Comp-LoRA (Ours) & \underline{69.97} & \underline{70.09} & {29.81} & \textbf{79.93} & \underline{69.84} & 84.40 & {91.62} & \textbf{85.38} & \textbf{94.52} & \textbf{59.16} & \textbf{77.72} & \textbf{73.85} \\
\midrule
\midrule
\multirow{11}{*}{\textbf{2}}
& CoOp  {\tiny \textbf{(IJCV '22)}}  & 68.7 & 68.0 & 28.1 & 66.2 & 70.5 & 82.6 & 89.9 & 80.9 & 93.0 & 53.7 & 73.5 & 70.5 \\
& CoCoOp {\tiny \textbf{(CVPR '22)}} & 70.1 & 69.4 & 29.3 & 61.8 & 68.4 & 85.9 &  \underline{91.9} & 77.8 & 94.4 & 52.3 & 73.4 & 70.4 \\
& TIP-Adapter-F {\tiny \textbf{(ECCV '22)}} & 70.0 & 68.6 & {32.8} & 73.2 & 70.8 & 86.0 & 91.6 & {90.1} & 93.9 & {57.8} & 76.2 & 73.7 \\
& CLIP-Adapter {\tiny \textbf{(IJCV '23)}} & 68.2 & 67.2 & 27.0 & 51.2 & 66.6 & 86.2 & 89.7 & 71.7 & 93.4 & 45.4 & 68.4 & 66.8 \\
& PLOT++ {\tiny \textbf{(ICLR '23)}} & 68.3 & 68.1 & 31.1 & {76.8} & \underline{73.2} & 86.3 & \textbf{92.3} & \underline{89.8} & {94.7} & 56.7 & {76.8} & {74.0} \\
& KgCoOp {\tiny \textbf{(CVPR '23)}} & 69.6 & 69.6 & 28.0 & 69.2 & 68.2 & \textbf{86.6} & \textbf{92.3} & 79.8 & 94.5 & 55.3 & 74.6 & 71.6 \\
& TaskRes {\tiny \textbf{(CVPR '23)}} & {70.2} & 70.5 & 32.7 & 70.2 & 72.1 & 85.6 & 90.7 & 84.4 & 94.3 & 55.6 & 75.2 & 72.9 \\
& MaPLe {\tiny \textbf{(CVPR '23)}} & 70.0 & {70.7} & 29.5 & 59.4 & 68.5 & \underline{86.5} & 91.8 & 79.8 & \underline{94.9} & 50.6 & 74.0 & 70.5  \\

& ProGrad {\tiny \textbf{(ICCV '23)}} & 69.1 & 69.0 & 31.1 & 66.3 & {72.4} & 84.8 & 91.5 & 87.5 & 93.6 & 56.0 & 75.6 & 72.4 \\
& APE {\tiny \textbf{(ICCV '23)}} & 70.60 &71.08 &34.08 &68.75 &70.56 &86.25 &90.81 &91.03 &\textbf{95.13} &\underline{60.17} &74.76 &73.93 \\
& CLIP-LoRA {\tiny \textbf{(CVPRW '24)}}& \textbf{70.8} & \underline{71.3} & \underline{33.2} & \underline{82.7} & \underline{73.2} & 83.2 & 91.3 & \underline{89.8} & 94.6 & {59.9} & \textbf{80.0} & \underline{75.5}  \\
\rowcolor{LightGray} & Comp-LoRA (Ours) & \underline{70.65}& \textbf{71.67}& \textbf{35.99}& \textbf{83.77}& \textbf{73.35}& 84.07& 91.39& \textbf{90.82}& 94.85& \textbf{61.15} & \underline{79.34} & \textbf{76.09} \\
\midrule
\midrule
\multirow{11}{*}{\textbf{4}}
& CoOp  {\tiny \textbf{(IJCV '22)}} &  69.7 & 70.6 & 29.7 & 65.8 & 73.4 & 83.5 & 92.3 & 86.6 & 94.5 & 58.5 & 78.1 & 73.0 \\
& CoCoOp {\tiny \textbf{(CVPR '22)}} & 70.6 & 70.4 & 30.6 & 61.7 & 69.5 & 86.3 & \underline{92.7} & 81.5 & 94.8 & 55.7 & 75.3 & 71.7 \\
& TIP-Adapter-F {\tiny \textbf{(ECCV '22)}} & 70.7 & 70.8 & {35.7} & 76.8 & 74.1 & 86.5 & 91.9 & 92.1 & 94.8 & 59.8 & 78.1 & 75.6 \\
& CLIP-Adapter {\tiny \textbf{(IJCV '23)}} & 68.6 & 68.0 & 27.9 & 51.2 & 67.5 & 86.5 & 90.8 & 73.1 & 94.0 & 46.1 & 70.6 & 67.7 \\
& PLOT++ {\tiny \textbf{(ICLR '23)}} & 70.4 & 71.7 & 35.3 & {83.2} & {76.3} & 86.5 & 92.6 & {92.9} & {95.1} & {62.4} & {79.8} & {76.9} \\
& KgCoOp {\tiny \textbf{(CVPR '23)}} & 69.9 & 71.5 & 32.2 & 71.8 & 69.5 & \textbf{86.9} & 92.6 & 87.0 & 95.0 & 58.7 & 77.6 & 73.9 \\
& TaskRes {\tiny \textbf{(CVPR '23)}} & \underline{71.0} &{72.7} & 33.4 & 74.2 & 76.0 & 86.0 & 91.9 & 85.0 & 95.0 & 60.1 & 76.2 & 74.7 \\
& MaPLe {\tiny \textbf{(CVPR '23)}} & 70.6 & 71.4 & 30.1 & 69.9 & 70.1 & \underline{86.7} & \textbf{93.3} & 84.9 & 95.0 & 59.0 & 77.1 & 73.5  \\
& ProGrad {\tiny \textbf{(ICCV '23)}} & 70.2 & 71.7 & 34.1 & 69.6 & 75.0 & 85.4 & 92.1 & 91.1 & 94.4 & 59.7 & 77.9 & 74.7 \\
& APE {\tiny \textbf{(ICCV '23)}} & 70.80& 72.36& 34.68& 75.77& 73.36& 86.27& 91.58& 94.64& 95.58& 65.54& 78.85   & 76.31 \\
& CLIP-LoRA {\tiny \textbf{(CVPRW '24)}}& \textbf{71.4} &  \underline{72.8} &  \underline{37.9} & \underline{84.9} & \textbf{77.4} & 82.7 & 91.0 & \underline{93.7} & \underline{95.2} & \underline{63.8} & \textbf{81.1} & \underline{77.4} \\
\rowcolor{LightGray} & Comp-LoRA (Ours) & \textbf{71.4} & \textbf{73.11} & \textbf{38.32} & \textbf{86.4} & \underline{76.73} & 82.7 & 90.29 & \textbf{94.03} & \textbf{95.28} & \textbf{64.54} & \underline{80.97} & \textbf{77.61} \\
\midrule
\midrule
\multirow{11}{*}{\textbf{8}}
& CoOp  {\tiny \textbf{(IJCV '22)}} & 70.8 & 72.4 & 37.0 & 74.7 & 76.8 & 83.3 & 92.1 & 95.0 & 94.7 & 63.7 & 79.8 & 76.4 \\
& CoCoOp {\tiny \textbf{(CVPR '22)}} & 70.8 & 71.5 & 32.4 & 69.1 & 70.4 & \underline{87.0} & \textbf{93.3} & 86.3 & 94.9 & 60.1 & 75.9 & 73.8 \\
& TIP-Adapter-F {\tiny \textbf{(ECCV '22)}} & {71.7} & 73.5 & 39.5 & 81.3 & 78.3 & 86.9 & 91.8 & 94.3 & 95.2 & {66.7} & 82.0 & 78.3 \\
& CLIP-Adapter {\tiny \textbf{(IJCV '23)}} & 69.1 & 71.7 & 30.5 & 61.6 & 70.7 & 86.9 & 91.9 & 83.3 & 94.5 & 50.5 & 76.2 & 71.5 \\
& PLOT++ {\tiny \textbf{(ICLR '23)}} & 71.3 & 73.9 & {41.4} & {88.4} & {81.3} & 86.6 & 93.0 & 95.4 & {95.5} & 66.5 & {82.8} & {79.6} \\
& KgCoOp {\tiny \textbf{(CVPR '23)}} & 70.2 & 72.6 & 34.8 & 73.9 & 72.8 & \underline{87.0} & 93.0 & 91.5 & 95.1 & 65.6 & 80.0 & 76.0 \\
& TaskRes {\tiny \textbf{(CVPR '23)}} & \underline{72.3} & {74.6} & 40.3 & 77.5 & 79.6 & 86.4 & 92.0 & {96.0} & 95.3 & {66.7} & 81.6 & 78.4 \\
& MaPLe {\tiny \textbf{(CVPR '23)}} & 71.3 & 73.2 & 33.8 & 82.8 & 71.3 & \textbf{87.2} & \underline{93.1} & 90.5 & 95.1 & 63.0 & 79.5 & 76.4 \\

& ProGrad {\tiny \textbf{(ICCV '23)}} & 71.3 & 73.0 & 37.7 & 77.8 & 78.7 & 86.1 & 92.2 & 95.0 & 94.8 & 63.9 & 80.5 & 77.4 \\
& APE {\tiny \textbf{(ICCV '23)}} & 71.27 &73.43 &39.51 &75.25 &74.19 &86.59 &91.63 &95.09 &\textbf{95.66} &\textbf{70.04} &78.61 &77.39 \\
& CLIP-LoRA {\tiny \textbf{(CVPRW '24)}} &  \underline{72.3} & \underline{74.7} & \underline{45.7} & \underline{89.7} & \textbf{82.1} & 83.1 & 91.7 & \textbf{96.3} & \underline{95.6} & {67.5} & \textbf{84.1} & \underline{80.3}  \\
\rowcolor{LightGray} & Comp-LoRA (Ours) & \textbf{72.41} & \textbf{75.2} & \textbf{46.65} & \textbf{90.68} & \underline{81.92} & 83.42 & 92.35 & \underline{96.22} & {95.54} & \underline{69.15} & \underline{83.82} & \textbf{80.67} \\
\midrule
\midrule
\multirow{11}{*}{\textbf{16}}
& CoOp  {\tiny \textbf{(IJCV '22)}} &  71.5 & 74.6 & 40.1 & 83.5 & 79.1 & 85.1 & 92.4 & 96.4 & 95.5 & 69.2 & 81.9 & 79.0 \\
& CoCoOp {\tiny \textbf{(CVPR '22)}} & 71.1 & 72.6 & 33.3 & 73.6 & 72.3 & \textbf{87.4} & \underline{93.4} & 89.1 & 95.1 & 63.7 & 77.2 & 75.4 \\
& TIP-Adapter-F {\tiny \textbf{(ECCV '22)}} & {73.4} &  {76.0} & 44.6 & 85.9 & 82.3 & 86.8 & 92.6 & 96.2 & 95.7 & 70.8 & 83.9 & 80.7 \\
& CLIP-Adapter {\tiny \textbf{(IJCV '23)}} & 69.8 & 74.2 & 34.2 & 71.4 & 74.0 & 87.1 & 92.3 & 92.9 & 94.9 & 59.4 & 80.2 & 75.5  \\
& PLOT++ {\tiny \textbf{(ICLR '23)}} & 72.6 & {76.0} & {46.7} & {92.0} & {84.6} & 87.1 & \textbf{93.6} & {97.6} & {96.0} & 71.4 & {85.3} & {82.1} \\
& KgCoOp {\tiny \textbf{(CVPR '23)}} & 70.4 & 73.3 & 36.5 & 76.2 & 74.8 & \underline{87.2} & 93.2 & 93.4 & 95.2 & 68.7 & 81.7 & 77.3 \\
& TaskRes {\tiny \textbf{(CVPR '23)}} & 73.0 & \underline{76.1} & 44.9 & 82.7 & 83.5 & 86.9 & 92.4 & 97.5 & 95.8 & {71.5} & 84.0 & 80.8 \\
& MaPLe {\tiny \textbf{(CVPR '23)}} &  71.9 & 74.5 & 36.8 & 87.5 & 74.3 & \textbf{87.4} & 93.2 & 94.2 & 95.4 & 68.4 & 81.4 & 78.6 \\
& ProGrad {\tiny \textbf{(ICCV '23)}} &  72.1 & 75.1 & 43.0 & 83.6 & 82.9 & 85.8 & 92.8 & 96.6 & 95.9 & 68.8 & 82.7 & 79.9 \\
& APE {\tiny \textbf{(ICCV '23)}} & 71.48& 74.22 &42.63& 81.57& 77.19& 86.72& 92.01& 94.84& 95.38& 69.98& 80.76  & 78.79 \\
& CLIP-LoRA {\tiny \textbf{(CVPRW '24)}}& \underline{73.6} & \underline{76.1} & \underline{54.7} & \underline{92.1}  & \underline{86.3} & 84.2 & 92.4 & \textbf{98.0} & \underline{96.4} & \underline{72.0} & \textbf{86.7} & \underline{83.0} \\
\rowcolor{LightGray} & Comp-LoRA (Ours) & \textbf{73.72} & \textbf{76.52} & \textbf{56.53} & \textbf{93.25}  & \textbf{87.1} & 84.21 & 93.14 & \underline{97.97} & \textbf{96.80} & \textbf{72.12} & \underline{86.62} & \textbf{83.45} \\
\bottomrule

\end{tabular}}

\end{table*}

\section{Extra Generality Experiments}
We conducted more designed experiments on catastrophic forgetting problem as shown in \cref{tab:leave_one_out_sun,tab:leave_one_out_fgvc,tab:leave_one_out_eurosat,tab:leave_one_out_cars,tab:leave_one_out_food,tab:leave_one_out_pets,tab:leave_one_out_flowers,tab:leave_one_out_caltech,tab:leave_one_out_dtd,tab:leave_one_out_ucf}. We fine-tuned on one few shot support set through the proposed and the baseline methods, and then tested their reserved zero shot classification ability on other 10 few shot tasks. These results showed that our Comp-LoRA method surpassed the baseline method on the suppression of the catastrophic forgetting problem.

\begin{table*}[h!]
\caption{Designed experiments on catastrophic forgetting problem. We fine-tuned on one few shot support set through the proposed and the baseline methods, and then tested their reserved zero shot classification ability on other 10 few shot tasks. The average column was taken on the other 10 few shot tasks(without SUN). We set the backbone as ViT-B/16. The complementary subspace dimension in Comp-LoRA was set as 496. We averaged over 5 random seeds for the Top-1 accuracy values. Highest value is highlighted in \textbf{bold}.}
\label{tab:leave_one_out_sun}
\centering
\resizebox{0.9\textwidth}{!}{
\begin{tabular}{l|c|ccccccccccc}
\toprule
    Method & SUN & ImageNet  & Aircraft & EuroSAT & Cars & Food & Pets &  Flowers & Caltech & DTD & UCF & Average
\\ \midrule 
{CLIP} {\tiny \textbf{(ICML '21)}} & 62.6 & 66.7 & 24.7 & 47.5 & 65.3 & 86.1 & 89.1 & 71.4 & 92.9 & 43.6 & 66.7 & 65.40 \\
\midrule
    CLIP-LoRA {\tiny \textbf{(CVPRW '24)}}& {76.1} & {66.55} & {18.69} & {32.00} & \textbf{60.95} & 82.49 & {86.75} & \textbf{68.17} &  92.94 & {44.27} & \textbf{66.06} & {61.88} \\
Comp-LoRA (Ours) & \textbf{76.52} & \textbf{67.10} & \textbf{19.95} & \textbf{41.93} & {59.52} & \textbf{83.70} & \textbf{87.27} & {67.48} & \textbf{93.31} & \textbf{45.63} & {65.48} & \textbf{63.13} \\
\bottomrule
\end{tabular}}
\end{table*}

\begin{table*}[h!]
\caption{Designed experiments on catastrophic forgetting problem. We fine-tuned on one few shot support set through the proposed and the baseline methods, and then tested their reserved zero shot classification ability on other 10 few shot tasks. The average column was taken on the other 10 few shot tasks(without SUN). We set the backbone as ViT-B/16. The complementary subspace dimension in Comp-LoRA was set as 496. We averaged over 5 random seeds for the Top-1 accuracy values. Highest value is highlighted in \textbf{bold}.}
\label{tab:leave_one_out_fgvc}
\centering
\resizebox{0.9\textwidth}{!}{
\begin{tabular}{l|c|ccccccccccc}
\toprule
    Method & Aircraft  & ImageNet & SUN & EuroSAT & Cars & Food & Pets &  Flowers & Caltech & DTD & UCF & Average
\\ \midrule 
{CLIP} {\tiny \textbf{(ICML '21)}} & 24.7  & 66.7 & 62.6 & 47.5 & 65.3 & 86.1 & 89.1 & 71.4 & 92.9 & 43.6 & 66.7 & 69.19 \\
\midrule
    CLIP-LoRA {\tiny \textbf{(CVPRW '24)}}& {54.31} & {66.45} & {61.78} & 34.77 & 57.68 & 83.47 & 88.99 & 66.67 & 88.36 & 43.62 & 63.34  & {65.51} \\
Comp-LoRA (Ours) & \textbf{54.76} & \textbf{66.88} & \textbf{63.36} & \textbf{37.28} & \textbf{58.79} & \textbf{83.81} & \textbf{89.40} & \textbf{67.84} & \textbf{91.68} & \textbf{43.97} & \textbf{66.16} & \textbf{66.91} \\
\bottomrule
\end{tabular}}
\end{table*}

\begin{table*}[h!]
\caption{Designed experiments on catastrophic forgetting problem. We fine-tuned on one few shot support set through the proposed and the baseline methods, and then tested their reserved zero shot classification ability on other 10 few shot tasks. The average column was taken on the other 10 few shot tasks(without SUN). We set the backbone as ViT-B/16. The complementary subspace dimension in Comp-LoRA was set as 496. We averaged over 5 random seeds for the Top-1 accuracy values. Highest value is highlighted in \textbf{bold}.}
\label{tab:leave_one_out_eurosat}
\centering
\resizebox{0.9\textwidth}{!}{
\begin{tabular}{l|c|ccccccccccc}
\toprule
    Method & EuroSAT  & ImageNet & SUN & Aircraft  & Cars & Food & Pets &  Flowers & Caltech & DTD & UCF & Average
\\ \midrule 
{CLIP} {\tiny \textbf{(ICML '21)}} &  47.5 & 66.7 & 62.6 & 24.7 & 65.3 & 86.1 & 89.1 & 71.4 & 92.9 & 43.6 & 66.7 & 66.91 \\
\midrule
    CLIP-LoRA {\tiny \textbf{(CVPRW '24)}}& \textbf{91.67} & {67.97} & {63.10} & 22.83 & 65.32 & 85.04 & 87.84 & \textbf{69.10} & 93.27 & 47.52 & 66.11  & {66.81} \\
Comp-LoRA (Ours) & {91.14} & \textbf{68.15} & \textbf{64.38} & \textbf{22.89} & \textbf{65.97} & \textbf{85.18} & \textbf{88.23} & {68.37} & \textbf{93.67} & \textbf{49.29} & \textbf{67.41} & \textbf{67.35} \\
\bottomrule
\end{tabular}}
\end{table*}

\begin{table*}[h!]
\caption{Designed experiments on catastrophic forgetting problem. We fine-tuned on one few shot support set through the proposed and the baseline methods, and then tested their reserved zero shot classification ability on other 10 few shot tasks. The average column was taken on the other 10 few shot tasks(without SUN). We set the backbone as ViT-B/16. The complementary subspace dimension in Comp-LoRA was set as 496. We averaged over 5 random seeds for the Top-1 accuracy values. Highest value is highlighted in \textbf{bold}.}
\label{tab:leave_one_out_cars}
\centering
\resizebox{0.9\textwidth}{!}{
\begin{tabular}{l|c|ccccccccccc}
\toprule
    Method & Cars  & ImageNet & SUN & Aircraft  & EuroSAT  & Food & Pets &  Flowers & Caltech & DTD & UCF & Average
\\ \midrule 
{CLIP} {\tiny \textbf{(ICML '21)}} &  65.3  & 66.7 & 62.6 & 24.7 & 47.5 & 86.1 & 89.1 & 71.4 & 92.9 & 43.6 & 66.7 & 65.13 \\
\midrule
    CLIP-LoRA {\tiny \textbf{(CVPRW '24)}}& {83.55} & 68.30 & 62.36 & 22.08 & 31.22 & 84.14 & \textbf{89.40} & \textbf{68.66} & 91.76 & 44.62 & 62.81  & 62.53 \\
    Comp-LoRA (Ours) & \textbf{84.23} & \textbf{68.78} & \textbf{63.19} & \textbf{22.56} & \textbf{40.06} & \textbf{84.40} & {89.29} & {67.28} & \textbf{92.74} & \textbf{45.74} & \textbf{63.94} & \textbf{63.80} \\
\bottomrule
\end{tabular}}
\end{table*}

\begin{table*}[h!]
\caption{Designed experiments on catastrophic forgetting problem. We fine-tuned on one few shot support set through the proposed and the baseline methods, and then tested their reserved zero shot classification ability on other 10 few shot tasks. The average column was taken on the other 10 few shot tasks(without SUN). We set the backbone as ViT-B/16. The complementary subspace dimension in Comp-LoRA was set as 496. We averaged over 5 random seeds for the Top-1 accuracy values. Highest value is highlighted in \textbf{bold}.}
\label{tab:leave_one_out_food}
\centering
\resizebox{0.9\textwidth}{!}{
\begin{tabular}{l|c|ccccccccccc}
\toprule
    Method & Food  & ImageNet & SUN & Aircraft  & EuroSAT  & Cars & Pets &  Flowers & Caltech & DTD & UCF & Average
\\ \midrule 
{CLIP} {\tiny \textbf{(ICML '21)}} &  86.1  & 66.7 & 62.6 & 24.7 & 47.5 & 65.3 & 89.1 & 71.4 & 92.9 & 43.6 & 66.7 & 63.05 \\
\midrule
    CLIP-LoRA {\tiny \textbf{(CVPRW '24)}}& \textbf{85.02} & \textbf{67.40}  & 63.79 & 21.87 & 37.63 & 63.28 & \textbf{88.55} & 65.08 & 91.81 & 41.31 & 64.90  & 60.56 \\
    Comp-LoRA (Ours) & {84.70} & {67.12} & \textbf{63.93} & \textbf{21.54} & \textbf{35.96} & \textbf{63.49} & {87.93} & \textbf{66.71} & \textbf{93.14} & \textbf{41.43} & \textbf{65.42} & \textbf{60.66} \\
\bottomrule
\end{tabular}}
\end{table*}

\begin{table*}[h!]
\caption{Designed experiments on catastrophic forgetting problem. We fine-tuned on one few shot support set through the proposed and the baseline methods, and then tested their reserved zero shot classification ability on other 10 few shot tasks. The average column was taken on the other 10 few shot tasks(without SUN). We set the backbone as ViT-B/16. The complementary subspace dimension in Comp-LoRA was set as 496. We averaged over 5 random seeds for the Top-1 accuracy values. Highest value is highlighted in \textbf{bold}.}
\label{tab:leave_one_out_pets}
\centering
\resizebox{0.9\textwidth}{!}{
\begin{tabular}{l|c|ccccccccccc}
\toprule
    Method & Pets  & ImageNet & SUN & Aircraft  & EuroSAT  & Cars & Food &  Flowers & Caltech & DTD & UCF & Average
\\ \midrule 
{CLIP} {\tiny \textbf{(ICML '21)}} &  89.1  & 66.7 & 62.6 & 24.7 & 47.5 & 65.3 & 86.1 & 71.4 & 92.9 & 43.6 & 66.7 & 62.75 \\
\midrule
    CLIP-LoRA {\tiny \textbf{(CVPRW '24)}}& {91.77} & 63.50  & 62.12 & \textbf{23.13} & 13.14 & 61.75 & \textbf{83.38} & 67.40 & 92.41 & 39.01 & \textbf{63.26}  & 56.91 \\
    Comp-LoRA (Ours) & \textbf{92.61} & \textbf{64.33} & \textbf{63.04} & {22.08} & \textbf{20.44} & \textbf{62.38} & {82.87} & \textbf{67.93} & \textbf{92.94} & \textbf{42.61} & {62.94} & \textbf{58.15} \\
\bottomrule
\end{tabular}}
\end{table*}

\begin{table*}[h!]
\caption{Designed experiments on catastrophic forgetting problem. We fine-tuned on one few shot support set through the proposed and the baseline methods, and then tested their reserved zero shot classification ability on other 10 few shot tasks. The average column was taken on the other 10 few shot tasks(without SUN). We set the backbone as ViT-B/16. The complementary subspace dimension in Comp-LoRA was set as 496. We averaged over 5 random seeds for the Top-1 accuracy values. Highest value is highlighted in \textbf{bold}.}
\label{tab:leave_one_out_flowers}
\centering
\resizebox{0.9\textwidth}{!}{
\begin{tabular}{l|c|ccccccccccc}
\toprule
    Method &  Flowers & ImageNet & SUN & Aircraft  & EuroSAT  & Cars & Food &  Pets & Caltech & DTD & UCF & Average
\\ \midrule 
{CLIP} {\tiny \textbf{(ICML '21)}} &  71.4  & 66.7 & 62.6 & 24.7 & 47.5 & 65.3 & 86.1 & 89.1 & 92.9 & 43.6 & 66.7 & 64.52 \\
\midrule
    CLIP-LoRA {\tiny \textbf{(CVPRW '24)}}& {97.36} & 65.45  & 60.22 & 21.06 & 33.46 & 62.92 & 80.83 & 88.61 & 91.03 & 42.08 & 63.15  & 60.88 \\
    Comp-LoRA (Ours) & \textbf{97.93} & \textbf{65.88} & \textbf{61.16} & \textbf{21.78} & \textbf{36.77} & \textbf{63.49} & \textbf{82.27} & \textbf{89.42} & \textbf{91.60} & \textbf{43.03} & \textbf{65.29} & \textbf{62.07} \\
\bottomrule
\end{tabular}}
\end{table*}

\begin{table*}[h!]
\caption{Designed experiments on catastrophic forgetting problem. We fine-tuned on one few shot support set through the proposed and the baseline methods, and then tested their reserved zero shot classification ability on other 10 few shot tasks. The average column was taken on the other 10 few shot tasks(without SUN). We set the backbone as ViT-B/16. The complementary subspace dimension in Comp-LoRA was set as 496. We averaged over 5 random seeds for the Top-1 accuracy values. Highest value is highlighted in \textbf{bold}.}
\label{tab:leave_one_out_caltech}
\centering
\resizebox{0.9\textwidth}{!}{
\begin{tabular}{l|c|ccccccccccc}
\toprule
    Method &  Caltech & ImageNet & SUN & Aircraft  & EuroSAT  & Cars & Food &  Pets & Flowers & DTD & UCF & Average
\\ \midrule 
{CLIP} {\tiny \textbf{(ICML '21)}} &  92.9  & 66.7 & 62.6 & 24.7 & 47.5 & 65.3 & 86.1 & 89.1 & 71.4 & 43.6 & 66.7 & 64.52 \\
\midrule
    CLIP-LoRA {\tiny \textbf{(CVPRW '24)}}& {96.35} & 66.20  & 62.91 & 20.31 & \textbf{41.37} & \textbf{60.83} & 82.30 & \textbf{88.74} & \textbf{63.99} & 43.03 & 64.21  & 59.39 \\
    Comp-LoRA (Ours) & \textbf{96.67} & \textbf{66.65} & \textbf{63.65} & \textbf{20.46} & {39.83} & {60.69} & \textbf{82.84} & {88.61} & {63.58} & \textbf{43.26} & \textbf{64.47} & \textbf{59.40} \\
\bottomrule
\end{tabular}}
\end{table*}

\begin{table*}[h!]
\caption{Designed experiments on catastrophic forgetting problem. We fine-tuned on one few shot support set through the proposed and the baseline methods, and then tested their reserved zero shot classification ability on other 10 few shot tasks. The average column was taken on the other 10 few shot tasks(without SUN). We set the backbone as ViT-B/16. The complementary subspace dimension in Comp-LoRA was set as 496. We averaged over 5 random seeds for the Top-1 accuracy values. Highest value is highlighted in \textbf{bold}.}
\label{tab:leave_one_out_dtd}
\centering
\resizebox{0.9\textwidth}{!}{
\begin{tabular}{l|c|ccccccccccc}
\toprule
    Method &  DTD & ImageNet & SUN & Aircraft  & EuroSAT  & Cars & Food &  Pets & Flowers & Caltech & UCF & Average
\\ \midrule 
{CLIP} {\tiny \textbf{(ICML '21)}} &  43.6  & 66.7 & 62.6 & 24.7 & 47.5 & 65.3 & 86.1 & 89.1 & 71.4 & 92.9 & 66.7 & 67.30 \\
\midrule
    CLIP-LoRA {\tiny \textbf{(CVPRW '24)}}& {69.98} & 61.60  & 61.83 & \textbf{21.63} & 34.63 & 61.60 & 78.74 & \textbf{87.19} & 57.00 & 90.43 & 63.18  & 61.78 \\
    Comp-LoRA (Ours) & \textbf{72.16} & \textbf{62.90} & \textbf{62.79} & {20.70} & \textbf{43.72} & \textbf{62.52} & \textbf{81.58} & {85.80} & \textbf{58.91} & \textbf{90.71} & \textbf{64.76} & \textbf{63.44} \\
\bottomrule
\end{tabular}}
\end{table*}

\begin{table*}[h!]
\caption{Designed experiments on catastrophic forgetting problem. We fine-tuned on one few shot support set through the proposed and the baseline methods, and then tested their reserved zero shot classification ability on other 10 few shot tasks. The average column was taken on the other 10 few shot tasks(without SUN). We set the backbone as ViT-B/16. The complementary subspace dimension in Comp-LoRA was set as 496. We averaged over 5 random seeds for the Top-1 accuracy values. Highest value is highlighted in \textbf{bold}.}
\label{tab:leave_one_out_ucf}
\centering
\resizebox{0.9\textwidth}{!}{
\begin{tabular}{l|c|ccccccccccc}
\toprule
    Method &  UCF  & ImageNet & SUN & Aircraft  & EuroSAT  & Cars & Food &  Pets & Flowers & Caltech & DTD  & Average
\\ \midrule 
{CLIP} {\tiny \textbf{(ICML '21)}} &  66.7  & 66.7 & 62.6 & 24.7 & 47.5 & 65.3 & 86.1 & 89.1 & 71.4 & 92.9 & 43.6 & 64.99 \\
\midrule
    CLIP-LoRA {\tiny \textbf{(CVPRW '24)}}& {85.33} & 64.30  & 60.08 & 21.90 & 27.17 & \textbf{62.33} & 80.58 & 82.39 & 62.48 & 88.48 & \textbf{42.61} & 59.23 \\
    Comp-LoRA (Ours) & \textbf{86.39} & \textbf{65.00} & \textbf{60.47} & \textbf{22.86} & \textbf{35.48} & {61.21} & \textbf{82.27} & \textbf{85.17} & \textbf{64.47} & \textbf{90.51} & {41.61} & \textbf{60.90} \\
\bottomrule
\end{tabular}}
\end{table*}

\section{Complementary Dimension Experiments}
\begin{figure}[htb]
    \centering
    \includegraphics[width=.8\linewidth]{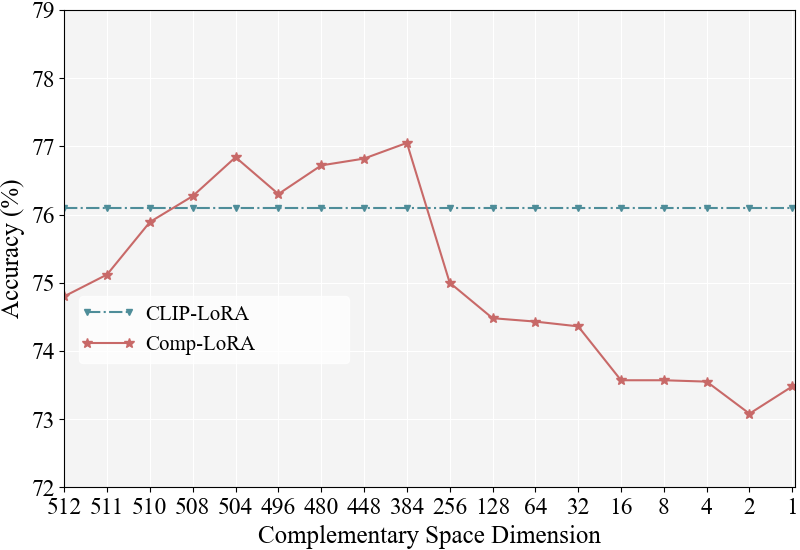}
    \caption{The effect of complementary subspace dimension on SUN397.
    }
    \label{fig:average_comp_rank_sun}
\end{figure}

\begin{figure}[htb]
    \centering
    \includegraphics[width=.8\linewidth]{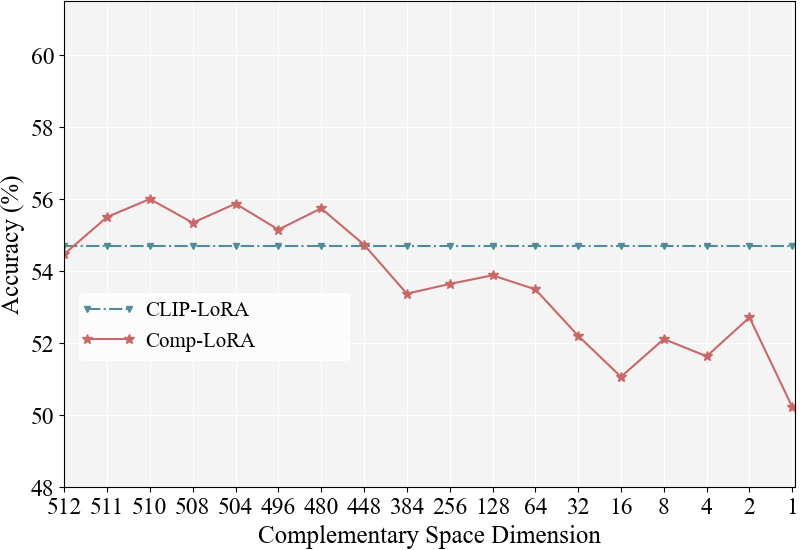}
    \caption{The effect of complementary subspace dimension on FGVC.
    }
    \label{fig:average_comp_rank_fgvc}
\end{figure}

\begin{figure}[htb]
    \centering
    \includegraphics[width=.8\linewidth]{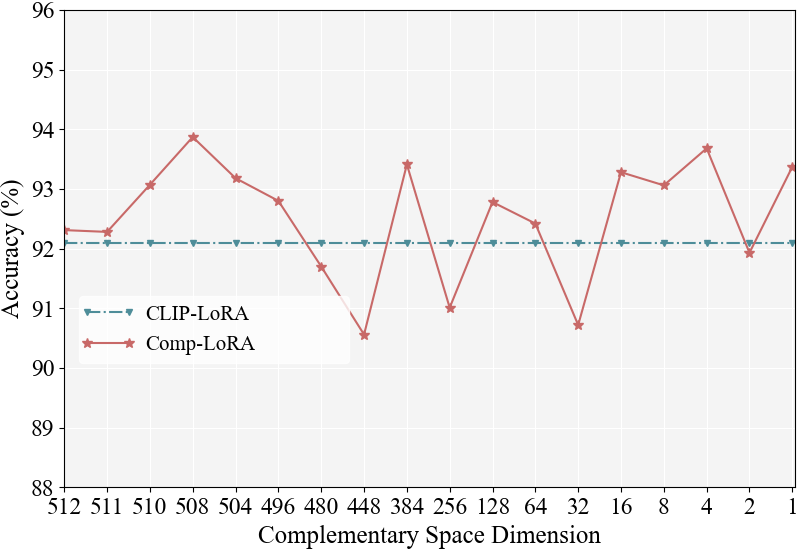}
    \caption{The effect of complementary subspace dimension on EuroSAT.
    }
    \label{fig:average_comp_rank_eurosat}
\end{figure}

\begin{figure}[htb]
    \centering
    \includegraphics[width=.8\linewidth]{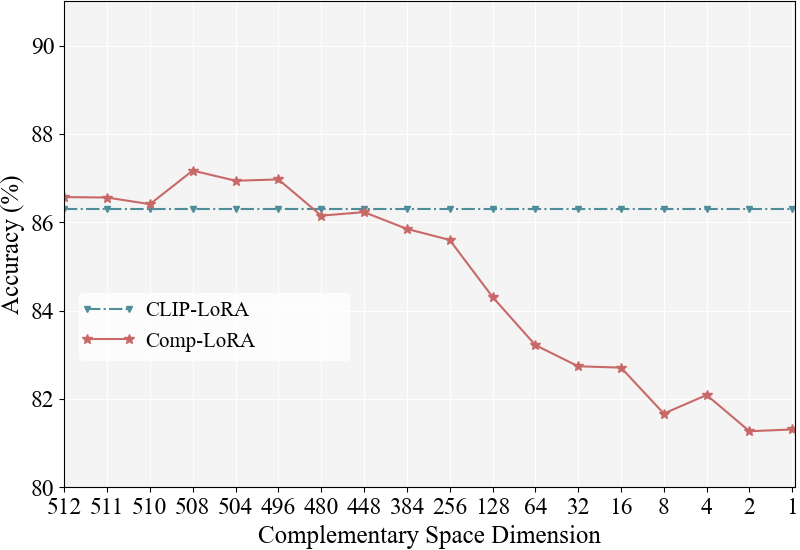}
    \caption{The effect of complementary subspace dimension on Stanford Cars.
    }
    \label{fig:average_comp_rank_cars}
\end{figure}

\begin{figure}[htb]
    \centering
    \includegraphics[width=.8\linewidth]{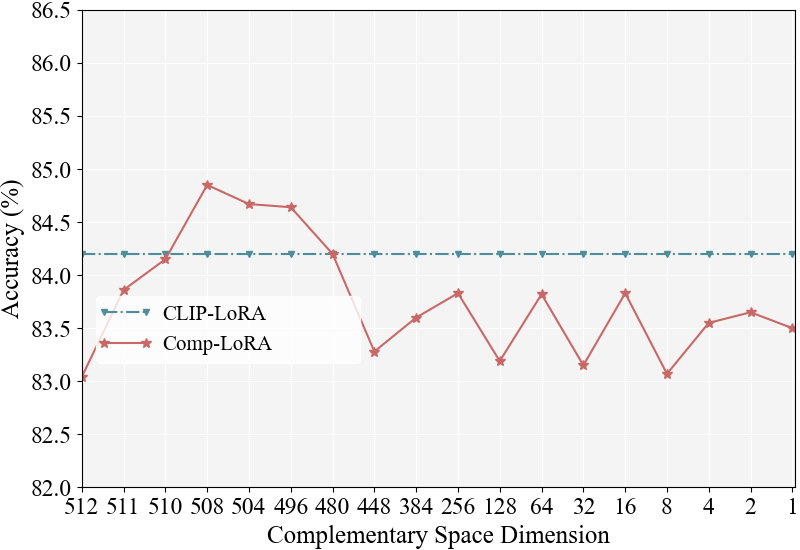}
    \caption{The effect of complementary subspace dimension on Food.
    }
    \label{fig:average_comp_rank_food}
\end{figure}

\begin{figure}[htb]
    \centering
    \includegraphics[width=.8\linewidth]{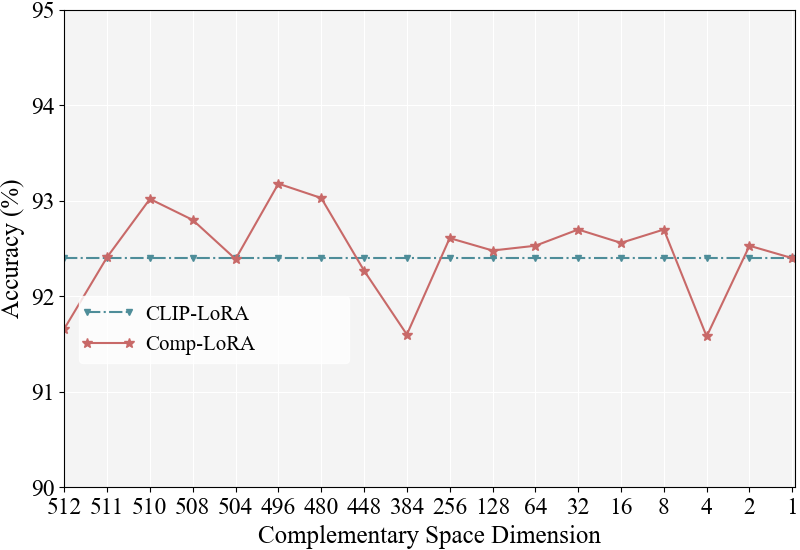}
    \caption{The effect of complementary subspace dimension on Pets.
    }
    \label{fig:average_comp_rank_pets}
\end{figure}

\begin{figure}[htb]
    \centering
    \includegraphics[width=.8\linewidth]{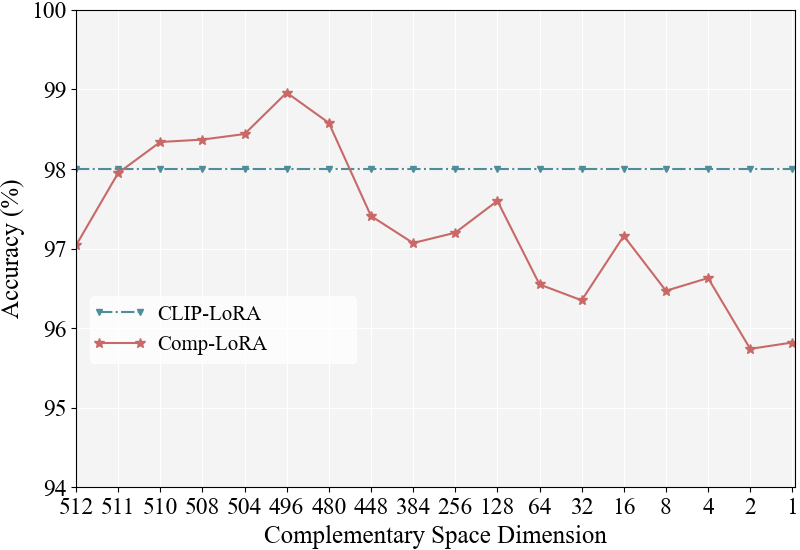}
    \caption{The effect of complementary subspace dimension on Flowers.
    }
    \label{fig:average_comp_rank_flowers}
\end{figure}

\begin{figure}[htb]
    \centering
    \includegraphics[width=.8\linewidth]{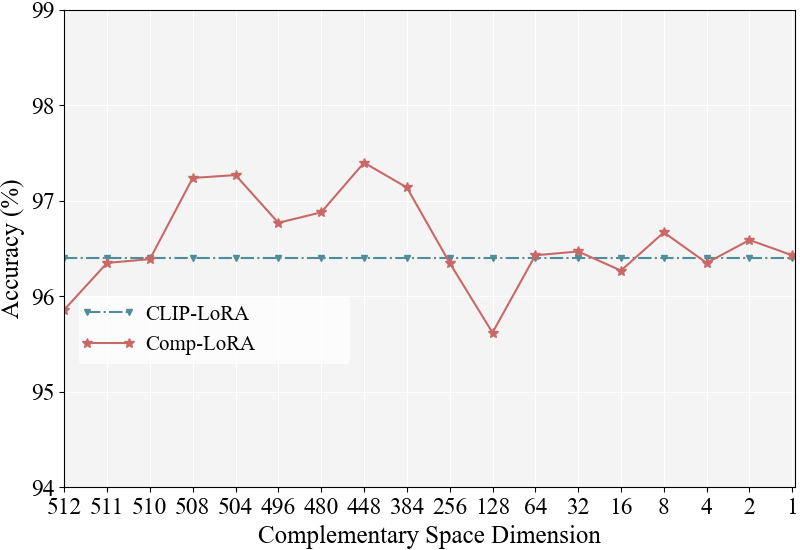}
    \caption{The effect of complementary subspace dimension on Caltech101.
    }
    \label{fig:average_comp_rank_caltech}
\end{figure}

\begin{figure}[htb]
    \centering
    \includegraphics[width=.8\linewidth]{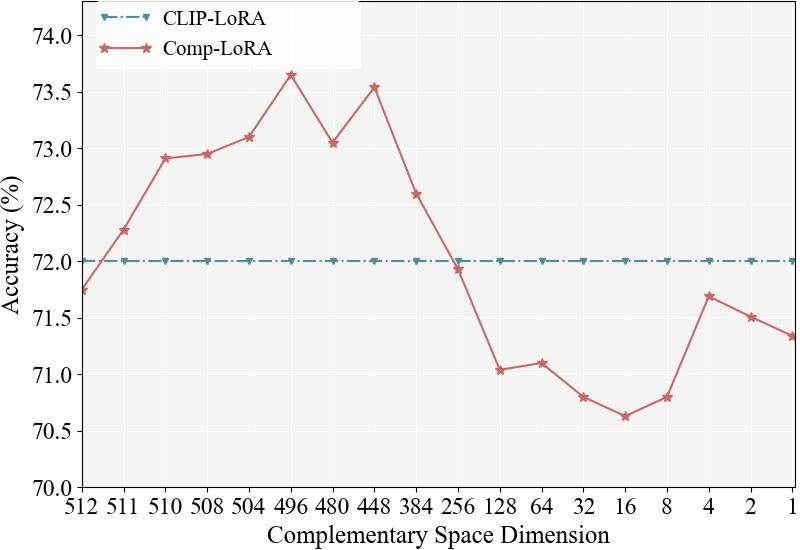}
    \caption{The effect of complementary subspace dimension on DTD.
    }
    \label{fig:average_comp_rank_dtd}
\end{figure}

\begin{figure}[htb]
    \centering
    \includegraphics[width=.8\linewidth]{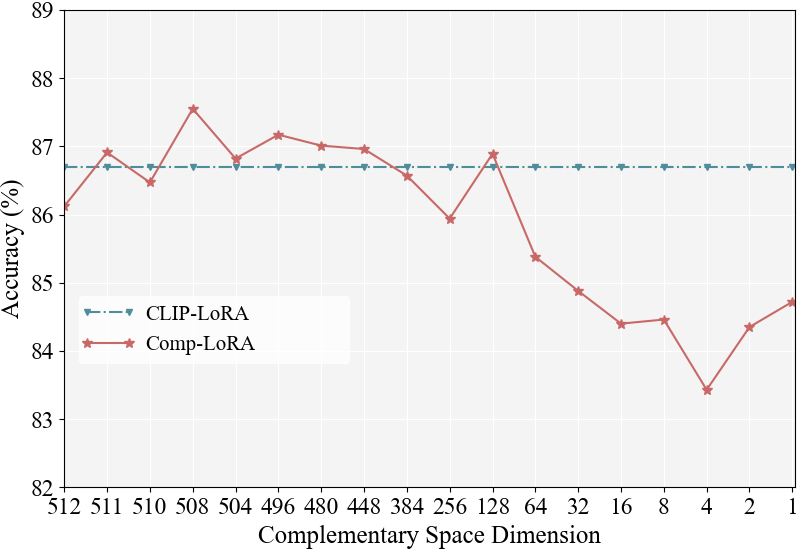}
    \caption{The effect of complementary subspace dimension on UCF101.
    }
    \label{fig:average_comp_rank_ucf}
\end{figure}

As shown in \cref{fig:average_comp_rank_sun,fig:average_comp_rank_fgvc,fig:average_comp_rank_eurosat,fig:average_comp_rank_cars,fig:average_comp_rank_food,fig:average_comp_rank_pets,fig:average_comp_rank_flowers,fig:average_comp_rank_caltech,fig:average_comp_rank_dtd,fig:average_comp_rank_ucf}, the univariate experiment on other experiments presents similar trend to that of ImageNet in \cref{fig:average_comp_rank}. In general, the performance of the proposed Comp-LoRA method increases at the beginning and then decreases when the complementary subspace dimension decreases. 
A few of these experiments demonstrated inconsistent trends compared with others. 
For example, the results on EuroSAT presents no obvious trend but oscillation. We doubt that depends on the specific dataset domain shift of EuroSAT. A future study may focus on this phenomenon. 

\end{document}